\documentclass[runningheads]{llncs}

\usepackage{eccv}
\usepackage{eccvabbrv}

\usepackage{microtype}
\usepackage{graphicx}
\usepackage{subcaption}
\usepackage{booktabs} 
\usepackage{multirow}
\usepackage{array}
\usepackage{adjustbox} 
\usepackage{pdflscape}
\usepackage{rotating}
\usepackage{array}
\usepackage{tcolorbox}

\usepackage{hyperref}

\usepackage{amssymb}
\usepackage{mathtools}

\usepackage[capitalize,noabbrev]{cleveref}

\ifodd 0
  \newcommand{\EditAni}[1]{\textcolor[rgb]{0,0,1}{#1}}
  \newcommand{\CommentAni}[1]{\textcolor[rgb]{1,0,0}{[Ani comment: #1]}}
  \newcommand{\CommentWu}[1]{\textcolor[rgb]{1,0,0}{[Wong comment: #1]}}
  \newcommand{\hlb}[1]{\textcolor{blue}{#1}}
  
\else
  \newcommand{\CommentAni}[1]{}
  \newcommand{\CommentWu}[1]{}  
  \newcommand{\EditAni}[1]{#1}
  \newcommand{\hlb}[1]{}
  
\fi

\usepackage[textsize=tiny]{todonotes}

\title{Understanding Semantic Perturbations on In-Processing Generative Image Watermarks}

\titlerunning{Understanding Semantic Perturbations on Generative Image Watermarks}

\begin{document}

\author{Anirudh Nakra \and Min Wu}

\authorrunning{Nakra and Wu}

\institute{University of Maryland, College Park, MD, USA}

\maketitle

\begin{abstract}
The widespread deployment of high-fidelity generative models has intensified the need for reliable mechanisms for provenance and content authentication. In-processing watermarking---embedding a signature into the generative model's synthesis procedure---has been advocated as a solution and is often reported to be robust to standard post-processing (such as geometric transforms and filtering). Yet robustness to semantic manipulations that alter high-level scene content while maintaining reasonable visual quality is not well studied or understood. We introduce a simple, multi-stage framework for systematically stress-testing in-processing generative watermarks under semantic drift. The framework utilizes off-the-shelf models for object detection, mask generation, and semantically guided inpainting or regeneration to produce controlled, meaning-altering edits with minimal perceptual degradation. Based on extensive experiments on representative schemes, we find that robustness varies significantly with the degree of semantic entanglement: methods by which watermarks remain detectable under a broad suite of conventional perturbations can fail under semantic edits, with watermark detectability in many cases dropping to near zero while image quality remains high. Overall, our results reveal a critical gap in current watermarking evaluations and suggest that watermark designs and benchmarking must explicitly account for robustness against semantic manipulation.
\keywords{Generative Image Watermarking \and Robust Watermarking \and Semantic Perturbations}
\end{abstract}

\section{Introduction}
\vspace{-4mm}

\begin{figure}[!t]
    \centering
    \includegraphics[width=\linewidth]{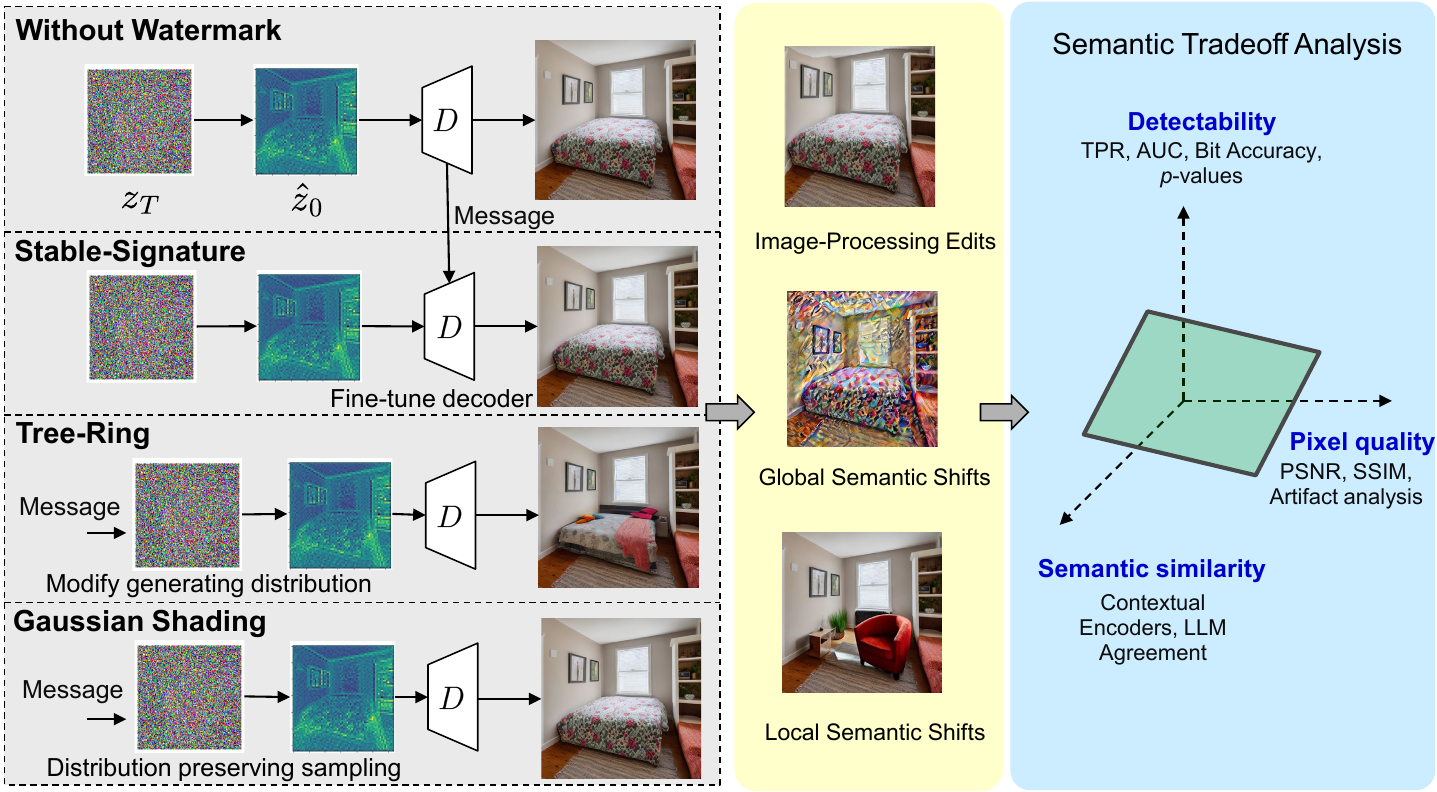}
    \caption{Representative in-processing generative image watermarks such as Stable-Signature~\cite{fernandez2023stable}, Tree-Ring~\cite{wen2023tree}, and Gaussian Shading~\cite{yang2024gaussian} aim to embed the watermarking signal during the LDM generation process. We study the resilience of these watermarks against a variety of semantic perturbations, aiming to quantify the level of semantic entanglement achieved by these methods.}
    \label{fig:main_overview}
    \vspace{-0.7cm}
\end{figure}

The rapid advancement and widespread accessibility of generative Artificial Intelligence (AI) have ushered in a new era of digital content creation. Models such as Stable Diffusion~\cite{rombach2022high} can produce photorealistic images from simple text prompts, facilitating creativity but also revealing new societal challenges. The ease with which synthetic media can be generated and disseminated has prompted an urgent need for reliable methods to establish data provenance—that is, to determine the origin of a piece of digital content. This is not merely an academic concern but a critical societal imperative for combating misinformation, protecting intellectual property, and ensuring the responsible deployment of AI technologies. This urgency is reflected in high-level policy initiatives, such as the U.S. Executive Order on Safe, Secure, and Trustworthy AI (Executive Order 14110)~\cite{whitehouse2023executive}, which explicitly calls for the development of standards and tools for content authentication and watermarking to label AI-generated content.

Digital watermarking for images has been advocated as a promising solution to address this provenance problem. A robust watermarking system is expected to possess several key properties: \textbf{imperceptibility}, ensuring the watermark does not degrade the visual quality of the content; \textbf{capacity}, allowing for the embedding of a sufficient amount of information; \textbf{robustness}, enabling the watermark's survival against common image manipulations;  and \textbf{security}, being resilient against adversarial attacks. Modern approaches have moved beyond simple post-processing techniques and instead integrate the watermark directly into the generative process of deep learning models, promising unprecedented levels of robustness \cite{wen2023tree, fernandez2023stable}.

However, the evaluation of this robustness has largely been confined to a specific class of pixel/sample-level perturbations. Existing benchmarks, such as WAVES~\cite{an2024waves}, provide a standardized and valuable framework for stress-testing watermarks against a diverse suite of attacks. Similarly, the original publications for representative methods such as Stable Signature and Tree-Ring demonstrate resilience against signal-level, or \textit{syntactic} perturbations such as JPEG compression, noise addition, geometric transforms, and various filtering operations \cite{wen2023tree, fernandez2023stable, an2024waves}. While these tests are essential, they may present a false sense of security because they do not account for a more potent and insidious threat vector for images in the GenAI era: attacks that target the \textit{semantic} content of images.

This paper's central thesis is that the very mechanism that makes modern generative watermarks robust—their deep integration into the semantic and latent fabric of the generative model—may also be their weakness. We argue that perturbations designed to manipulate specific \textit{meaning} of an image, rather than just its pixel values, may circumvent the defenses of these sophisticated watermarking schemes. To investigate this conjecture, we introduce a modular framework for launching targeted semantic perturbations as illustrated in Fig.~\ref{fig:main_overview}. This framework systematically identifies salient objects within an image and replaces them with new, semantically coherent content using state-of-the-art text-guided diffusion models. Overall, our contributions are threefold:

\vspace{1mm}
\noindent
$\bullet$ We present a multi-stage framework to study the entanglement of scene semantics with the watermarking process, which combines object detection with stylization methods and generative inpainting to perform content manipulation in a practical black-box setup.

\noindent
$\bullet$ We conduct an extensive empirical study on representative in-processing generative image watermarks, demonstrating that robustness varies sharply with the degree of semantic entanglement: methods by which remain detectable under a broad suite of conventional perturbations can fail under semantic edits. 

\noindent
$\bullet$ We provide a broader and stronger suite of image processing baseline perturbations on representative in-processing generative image watermarks.
    
Our findings in this paper challenge the prevailing assumptions about watermark robustness and highlight a critical new direction for the development and evaluation of data provenance technologies for generative AI.

\vspace{-3mm}
\section{Background, Notation, and Related Prior Art}
\vspace{-2mm}
\subsection{Latent Diffusion Models (LDM)}
\vspace{-1mm}
Diffusion models are probabilistic generative models that learn to transform samples from a simple prior distribution (typically Gaussian noise) into data samples via a sequence of learned denoising steps. They have achieved strong performance across image synthesis~\cite{rombach2022high}, super-resolution~\cite{gao2023implicit}, and related conditional generation tasks~\cite{song2025diffsim,demirag2024benchmarking}. As a result, latent diffusion models \cite{rombach2022high} have become a standard backbone for high-quality text-to-image generation. Unlike pixel-space diffusion, latent diffusion performs the forward noising and reverse denoising processes in a lower-dimensional latent space obtained via a learned encoder/decoder (such as a variational autoencoder (VAE)~\cite{kingma2013auto}). Operating in this compact representation improves computational efficiency and guides the model toward capturing higher-level semantic structure, since the latent space abstracts away some low-level pixel variability while preserving the global composition.

LDMs define a diffusion process in a learned latent space. Let $x \in \mathbb{R}^{H \times W \times 3}$ denote an image. An encoder $E$ maps $x$ to a latent representation
$z_0 = E(x) \in \mathbb{R}^{h \times w \times d},$ and a decoder $D$ maps latents back to pixel space, $\hat{x} = D(z_0)$. Using the denoising diffusion implicit models (DDIM) sampling strategy~\cite{song2020denoising}, the forward and inverse processes are as follows.

\paragraph{Forward (noising) process.}
Given a variance schedule $\{\beta_t\}_{t=1}^T$, define $\alpha_t = 1-\beta_t$ and $\bar{\alpha}_t = \prod_{s=1}^t \alpha_s$. The forward process corrupts $z_t$ as,

\begin{equation}
z_{t+1}=\sqrt{\alpha_{t+1}}\,z_t+\sqrt{1-\alpha_{t+1}}\,\varepsilon_{t+1},
\qquad \varepsilon_{t+1}\sim\mathcal{N}(0,I).
\end{equation}

\paragraph{Backward (denoising) process.}
Conditioned on $c$ (such as a text embedding), the denoiser $\varepsilon_\theta$ predicts the injected noise $\varepsilon_\theta(z_t,t,c) \approx \varepsilon$. This yields an estimate of the clean latent,
\begin{equation}
\hat{z}_0(z_t,t,c) = \frac{1}{\sqrt{\bar{\alpha}_t}}
\left(z_t - \sqrt{1-\bar{\alpha}_t}\,\varepsilon_\theta(z_t,t,c)\right).
\end{equation}

\vspace{-5mm}
\subsection{In-Processing Semantic Generative Image Watermarks}
\vspace{-2mm}
Existing generative image watermarking approaches can be broadly categorized into: (i) \textit{post-hoc watermarking}~\cite{tancik2020stegastamp,fernandez2022watermarking,ingemar2008digital,zhang2019robust}, which embeds a watermark after generation via a separate processing step, and (ii) \textit{in-processing watermarking}, which modifies the generation procedure ~\cite{fernandez2023stable,wen2023tree,yang2024gaussian} itself to produce watermarked outputs. Prior work often investigates post-hoc approaches under adversarial transformations, whereas investigations on in-processing methods that induce a more robust structured change in the generative distribution, such that the watermark can be recovered reliably from samples, often via higher-level statistical or semantic regularities, are lacking. In this paper, we focus on in-processing methods and analyze their vulnerabilities under structured semantic perturbations.

A prominent representative is \textbf{StableSignature} \cite{fernandez2023stable}, which embeds a watermark by fine-tuning the VAE decoder of a LDM. The decoder is conditioned on a binary signature, so that decoding latents through the fine-tuned module yields images that carry an imperceptible signature that can later be recovered by the corresponding detector. \textbf{Tree-Ring watermarking} \cite{wen2023tree} takes a conceptually different approach tailored to diffusion sampling. Rather than modifying pixels, Tree-Ring embeds a predefined pattern into the Fourier-domain representation of the initial noise (e.g., $z_T$ in latent diffusion). Under DDIM sampling, which is deterministic for a fixed conditioning and initialization, a verifier with access to the model can approximately invert the generation process for a candidate image to recover an estimate of the initial noise. Watermark presence is then tested by applying an FFT to the recovered noise and checking for the target Fourier pattern. Finally, \textbf{Gaussian Shading}~\cite{yang2024gaussian} is a message-conditioned semantic watermark that modulates latent sampling using a cryptographically generated bitstring. An encrypted message determines a sequence of partitioning decisions over the latent space; each bit selects a region from which the corresponding latent component is sampled. This construction is designed to preserve the global Gaussian prior while inducing a recoverable dependence between the sampled latent and the hidden message, which can subsequently be decoded from the generated image.

\vspace{-3mm}
\subsection{Semantic Evaluation of Generative Image Watermarks}
\vspace{-2mm}

While several works recognize that generative watermarks may be coupled to high-level image structure, none offer a comprehensive evaluation of in-processing watermarks under semantic manipulation. Lu et al.’s W-Bench~\cite{lu2024robust} studies post-hoc watermarks under image-editing operations, but does not explicitly model semantic drift beyond the edit areas, and its conclusions do not directly transfer to in-processing settings where watermarking alters the generative distribution and pixel-level imperceptibility is not the central objective. Arabi et al.’s SEAL~\cite{arabi2025seal} introduces a semantic “cat attack,” but evaluates under a forgery threat model without systematically varying semantic change. Closest in spirit to our work, Tallam et al.~\cite{tallam2025removing} employ a detect-and-replace pipeline against Tree-Ring, yet their procedure regenerates most of the background while preserving scene semantics, effectively approximating re-synthesis of the image, and without quantifying drift or semantic entanglement. Lukovnikov et al.~\cite{lukovnikovsemantic} consider a specific aspect of semantics, the layout control, via ControlNet~\cite{zhang2023adding} and find limited/no impact on watermarking, and their findings do not apply to other types of semantic shifts. 
TAG-WM~\cite{chen2025tag} explicitly considers semantic manipulation as an adversarial strategy and includes such edits in their experimental suites. It primarily considers semantics-preserving edits and does not quantify robustness as a function of semantic drift. Also, it studies specific tampering threat models and relies on conventional fidelity metrics, which do not characterize the degree to which the watermark is entangled with image semantics. 
In contrast, we directly manipulate semantic content as a function of semantic drift and quantify the resulting watermark robustness, addressing a gap in prior evaluation protocols.
\vspace{-5mm}
\section{Problem Formalization: Black-Box Threat Model}
\vspace{-2mm}
Let $\mathcal{G}$ be a generative model that produces a watermarked image $x_w \in \mathcal{X}$ from a latent code $z$ and a signature $s$: $x_w = \mathcal{G}(z, s)$.
Let $\mathcal{D}$ be a watermark detector/decoder that, for an image input, outputs a detection statistic and/or a decoded signature.
Given $x_w$, an adversary $\mathcal{A}$ applies a transformation $f:\mathcal{X}\rightarrow\mathcal{X}$ to obtain $x' = f(x_w)$ that reduces watermark detectability while maintaining perceptual fidelity.

\noindent\textbf{Semantic Perturbation Operator.} We model semantic editing as masked regeneration with an inpainting operator. Let $M\in\{0,1\}^{H\times W}$ be a binary mask identifying a target region (such as an object instance) and let $c$ denote a target semantic concept (such as a text prompt). We define the semantic perturbation operator $\mathcal{P}$:
\begin{equation}
\mathcal{P}_{\mathrm{sem}}(x_w, M, c)
= (1-M)\odot x_w \;+\; M\odot \mathrm{Inpaint}(x_w, M, c),
\end{equation}
where $\odot$ denotes the pixel-wise dot product and $\mathrm{Inpaint}(\cdot)$ is an editing tool (such as Stable Diffusion inpainting).

In the black-box setting, the adversary does not know the signature $s$ and has no access to the parameters of $\mathcal{G}$ or $\mathcal{D}$. The adversary is given only the watermarked image $x_w$ and access to public semantic editing tools (e.g., off-the-shelf diffusion inpainting models). The adversary produces an edited image $x'=\mathcal{P}_{\mathrm{sem}}(x_w, M, c)$, by selecting a mask $M$ and concept $c$ (e.g., via an object detector/segmenter and a prompt selection rule). This emulates the paradigm of Model-As-A-Service (MaaS), where users may only have query-level access to the model and can synthesize new generations using the service provider's model, but do not interact with it directly and do not know which watermarking scheme was used.

\begin{figure}[!t]
  \centering

  \begin{subfigure}[t]{0.44\linewidth}
    \centering
    \includegraphics[width=\linewidth]{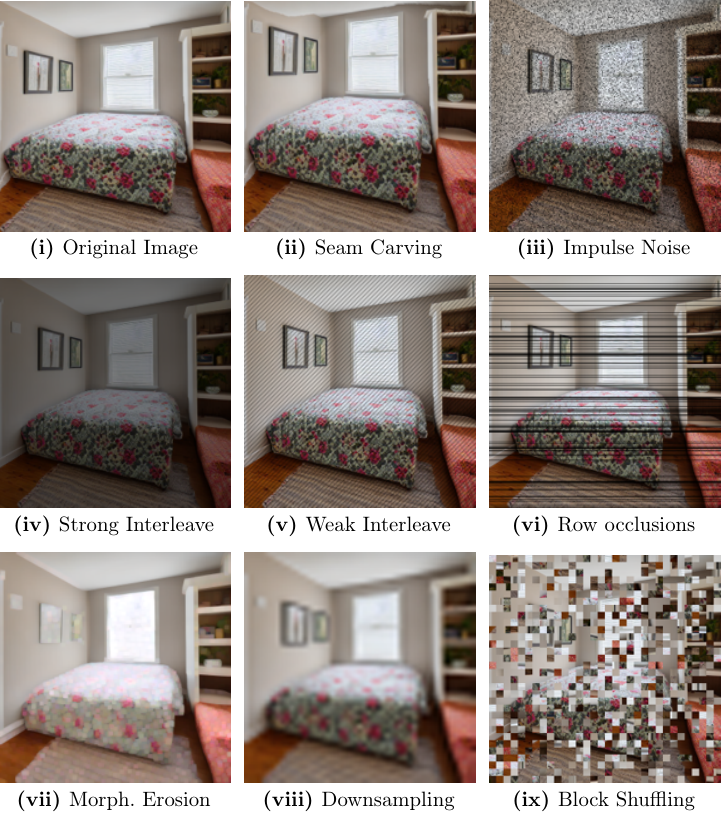}
    \caption{Image-processing-based perturbations.}
    \label{fig:3x3grid}
  \end{subfigure}\hfill\vrule\hfill
  \begin{subfigure}[t]{0.55\linewidth}
    \centering
    \includegraphics[width=\linewidth]{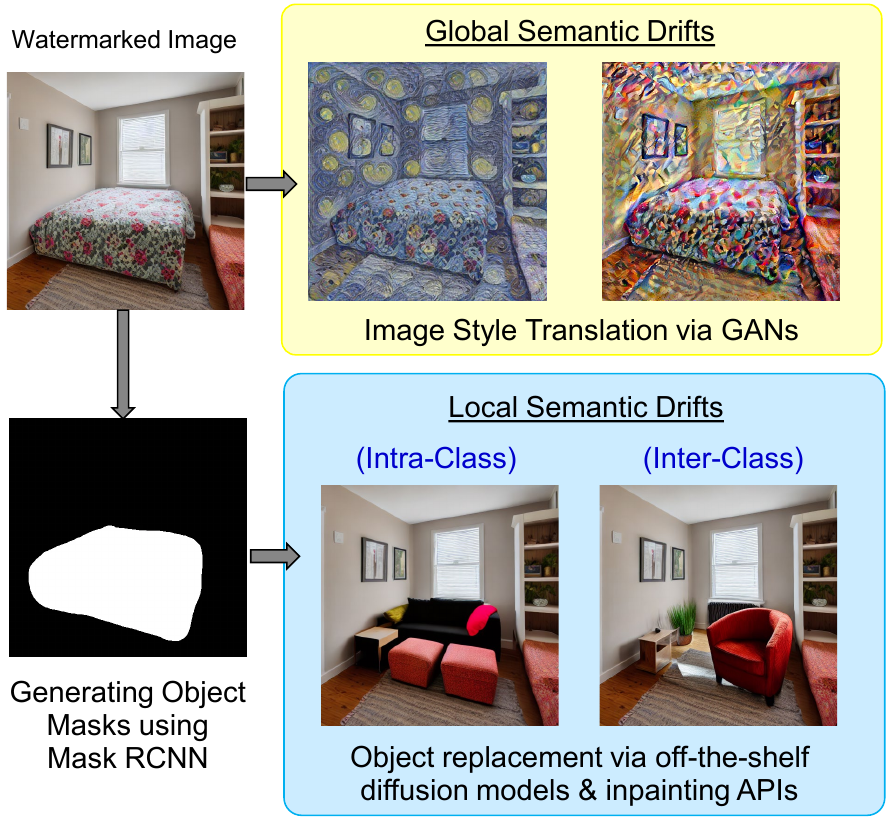}
    \caption{Semantic perturbations.}
    \label{fig:perturbation_framework}
  \end{subfigure}

  \caption{Overview of evaluation: (a)~Examples of different improved image-processing-based perturbations. (b)~The modular pipeline used to generate global \& local semantic perturbations.}
  \label{fig:pixel_and_pipeline}
  \vspace{-6mm}
\end{figure}

\vspace{-3mm}
\section{Improved Baselines: Image-Processing based Perturbations} 
\label{sec:image_processing}
\vspace{-3mm}
Current watermarking methods are typically evaluated under generic channel degradations—such as additive Gaussian noise, JPEG compression, and global brightness/contrast adjustments—but largely omit more structured, semantics-aware perturbations. Moving beyond this evaluation regime, we stress-test representative in-processing watermarking schemes under a suite of stronger, more targeted perturbations (illustrated in Fig.~\ref{fig:pixel_and_pipeline}(a)) that operate beyond simple low-level transformations.

\vspace{1mm}
\noindent\textbf{Traditional Content Aware Resizing.} Seam carving~\cite{avidan2023seam} is a content-aware image resizing technique that changes the size of an image with possibly different aspect ratios by iteratively removing low-energy seams (or inserting seams to enlarge the image). It is widely deployed in consumer editing tools (such as Adobe Photoshop and GIMP~\cite{GIMP}), and a post-processing operation that watermarked images may undergo in practice. Accordingly, we evaluate the robustness of generative watermarking schemes under seam-carving-based resizing.

\vspace{1mm}
\noindent\textbf{Image Downsampling.} Many GenAI watermarking methods~\cite{wen2023tree,alam2025specguard,kassis2025unmarker} embed signals in the spectral domain. Naïve image downsampling without appropriate low-pass filtering can introduce aliasing artifacts that distort frequency-domain amplitude and phase, potentially corrupting embedded watermark structure. Motivated by this, we evaluate watermark robustness under a downsample--upsample pipeline, measuring detectability after reducing resolution and subsequently restoring the image to its original size.

\vspace{1mm}
\noindent\textbf{Morphological Image Filtering.} Prior work has primarily considered simple photometric edits (e.g., contrast adjustments) and largely omits more structured image-processing operations. Morphological transformations~\cite{gonzalez2009digital}---including erosion and dilation---are routinely used for shape-based analysis and as post-processing steps in denoising and cleanup pipelines. While not inherently adversarial, they are commonly applied in practice and can alter local geometry and edge structure in ways that may affect watermark statistics. Accordingly, we evaluate the robustness of watermarked images under standard morphological operations.

\vspace{1mm}
\noindent\textbf{Impulse Noise/ Bit Flip and block shuffling.} We investigate the robustness of watermarks to perturbations that reduce image fidelity without fundamentally altering the image’s semantic content. We consider both stochastic and structured distortions. For stochastic noise, we test impulse corruption (pixel erasure), in which each pixel is independently lost/corrupted with probability $p$. For structured distortions that mimic synchronization errors in transmission or storage, we evaluate spatial desynchronization via pixel permutations and block-wise permutations, which preserve global content statistics but disrupt local spatial coherence and alignment.

\vspace{-4mm}
\section{Semantically Corrupting GenAI Watermarked Images}
\vspace{-3mm}

\begin{figure*}[!t]
\centering
\includegraphics[width=\linewidth]{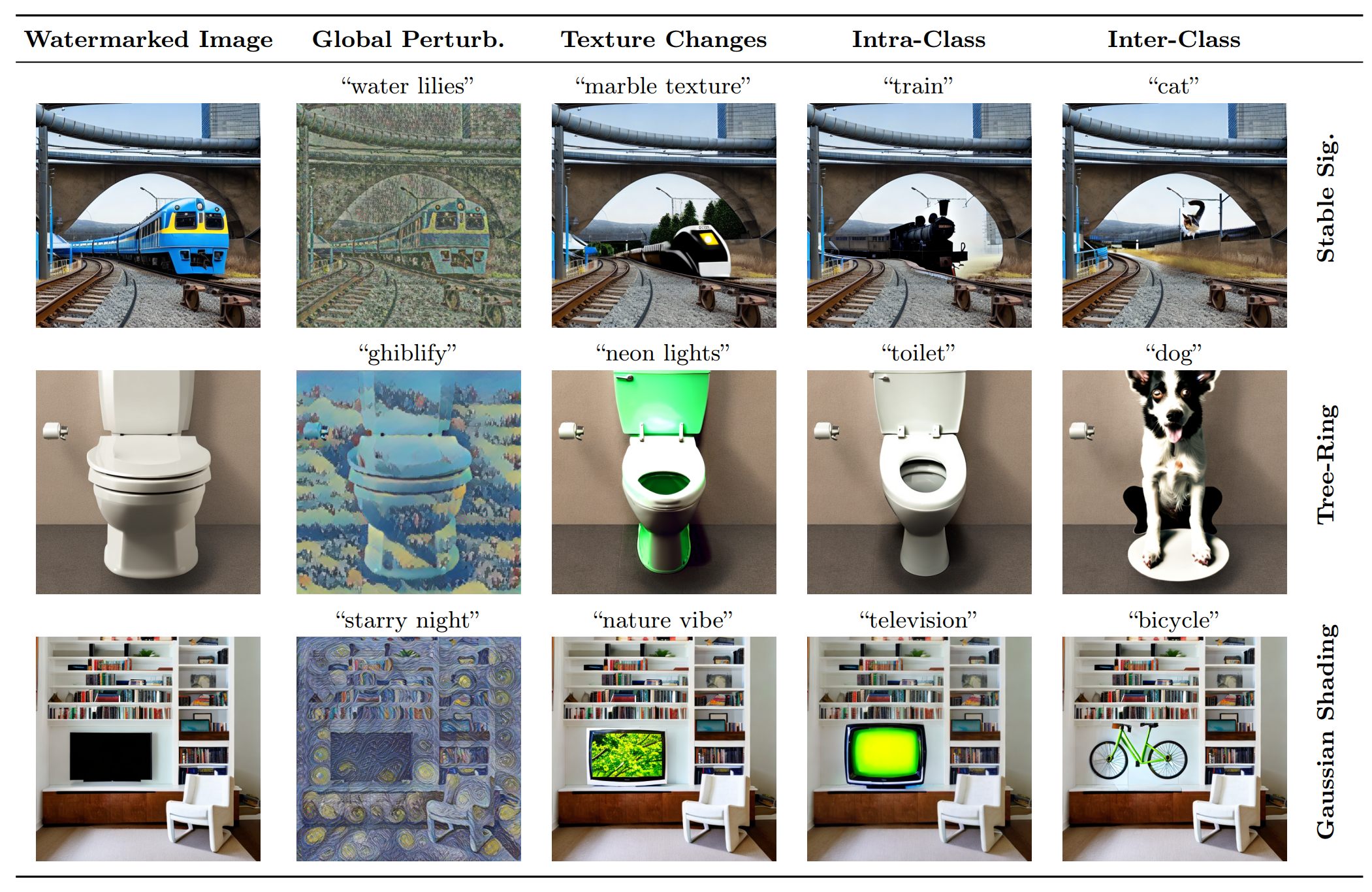}
\caption{An overview of semantic perturbations across representative in-processing watermarking methods.}
\label{fig:qualitative_watermark_perturbation}
\vspace{-0.7cm}
\end{figure*}

To probe the robustness of semantically embedded watermarks, we present a modular, multi-stage perturbation framework illustrated in Fig.~\ref{fig:pixel_and_pipeline}(b). Rather than relying on additive noise or low-level image transformations, the framework performs targeted, semantically informed content replacement. Concretely, we instantiate a detect-and-replace pipeline that composes off-the-shelf deep learning models to (i) localize semantically meaningful regions and (ii) replace or regenerate their content in a controlled manner. We illustrate qualitative results of our pipeline in Fig.~\ref{fig:qualitative_watermark_perturbation}.
\vspace{-3mm}
\subsection{Stage 1: Semantic Target Identification}
\vspace{-2mm}
The first stage identifies a semantically meaningful region within a watermarked image that will be manipulated. We use Mask R-CNN for instance segmentation \cite{he2017mask}, which produces pixel-accurate masks for each detected object instance rather than coarse bounding boxes. This granularity is critical for the subsequent inpainting step, as it cleanly removes the target object and yields a well-defined region to be regenerated. In each image, we enumerate all detected foreground instances and select the most salient object, in terms of the instance with the largest mask area. We then remove (mask out) this region, producing a binary mask that defines the target for semantic replacement.
\vspace{-3mm}
\subsection{Stage 2: Content Corruption and Regeneration}
\vspace{-2mm}

Given the target mask, the second stage replaces the removed region with semantically plausible content. The objective is to synthesize pixels that are consistent with the surrounding context while being sufficiently different in structure and texture to interfere with the watermark signal.

We instantiate this step using text-guided diffusion inpainting (such as Stable Diffusion), which conditions generation on the original image, the binary mask, and an edited text prompt. This formulation provides direct control over the semantics of the regenerated region and enables a continuum of perturbations, ranging from mild attribute edits (such as ``a bed with floral patterns”) to full object substitution (such as replacing ``a bed” with ``a sofa”). This controllability is central to our study, as it allows us to systematically vary semantic content while holding the global image context fixed, thereby isolating the sensitivity of generative watermarks to semantic manipulation.

\noindent\textbf{Local Edits.} To quantify the watermark's sensitivity to varying degrees of semantic change, we categorize local perturbations into three tiers based on the ``semantic gap" introduced during regeneration:

\noindent$\bullet$ \textbf{Intra-Class Texture Shift:} We modify only the textural attributes of the identified object (e.g., changing a ``plain shirt" to a ``flannel shirt"). This tests if the watermark is tied to specific high-frequency textural features within the latent embedding. 

\noindent$\bullet$ \textbf{Intra-Class Replacement:} We replace the object with a different instance of the same semantic class (e.g., replacing one ``chair'' with another ``chair'' via inpainting). This preserves the class-level semantics but alters the local geometry and fine-grained latent structure. 

\noindent$\bullet$ \textbf{Inter-Class Substitution:} We substitute the original object with an entirely different semantic category (e.g., ``dog'' to ``fire hydrant''). This represents the maximum semantic displacement possible within a local region while maintaining scene plausibility. 

\noindent\textbf{Global Semantic Shifts.} Beyond local substitutions, we evaluate the impact of global modifications that alter the entire image manifold. We implement this through neural style transfer (NST) \cite{gatys2016image}, which redistributes the image's low-level statistics to match a target style (such as transforming a photorealistic image into an oil painting). Unlike local edits, this shift affects every pixel simultaneously allowing us to compare whether generative watermarks are more resilient to global textural shifts than to targeted, local identity changes.

\vspace{-4mm}
\section{Evaluation}
\vspace{-2mm}
In this section, we conduct a comprehensive experimental evaluation to compare the resilience of representative in-processing watermarks against improved image-processing baselines and under our proposed semantic perturbation framework.
\vspace{-8mm}
\subsection{Experimental Setup}
\vspace{-2mm}
\textbf{Datasets.} We use the MS-COCO 2017~\cite{lin2014microsoft} validation split, comprising 5,000 captioned images, and resize all images to $512\times512$. For the three representative in-processing watermarking methods (namely, StableSignature, Tree-Ring, and Gaussian Shading), we use the corresponding MS-COCO captions as prompts to generate images through an LDM. 

\vspace{1mm}
\noindent \textbf{Watermark Generation.} For all three in-processing watermarking methods, we use Stable Diffusion v2-1~\cite{rombach2022high} as the base latent diffusion model. Given a prompt, we generate clean (unwatermarked) images by sampling the initial latent from the model’s standard prior. For Tree-Ring, we inject the watermark by adding concentric circular patterns in the frequency-domain representation of the initial latents, using the default hyperparameters. For Gaussian Shading, we sample latents conditioned on an encrypted per-image message generated using a randomly sampled key and nonce; we adopt the default configuration with a 256-bit message capacity and $\times 64$ message replication. For StableSignature, we use the official implementation with a 48-bit signature, and we decode images using the Stable Diffusion v2 decoder.

\vspace{1mm}
\noindent\textbf{Image-Processing-based Pixel Perturbation Setup.} We use backward energy-based seam carving~\cite{huang2009real} to iteratively remove 10--50\% of seams from the width of the watermarked image. For morphological operations, we examine erosion and dilation kernels of various sizes from $3\times3$ to $11\times11$. When analyzing erasure noise, we examine different probabilities of bit drop of 0--0.5. Finally, for understanding resilience to block shuffling, we iteratively change the size of the block being shuffled for 0--10\% of the image and the number of blocks that are permuted from 0--50\% of total blocks.

\vspace{1mm}
\noindent\textbf{ML-based Semantic Perturbation Setup.}  We use a pretrained Mask R-CNN with a ResNet-101 backbone (trained on MS-COCO) to obtain instance masks, which serve as the segmentation backbone for all local semantic edits. For mask-conditional generation, we employ Stable Diffusion v2-1 finetuned on LAMA-based inpainting mode~\cite{rombach2022high} across local perturbation settings. Specifically, we augment the region-specific prompt with an inpainting instruction drawn from the Stable Diffusion Prompts dataset~\cite{wangDiffusionDBLargescalePrompt2022}, conditioning the added text on the perturbation mode and the intra-mask content. To induce global semantic shifts, we apply neural style transfer using a VGG backbone.

\vspace{1mm}
\noindent\textbf{Watermark Detectability Metrics.} We evaluate robustness to several pixel- and layout-level perturbations. 
To ensure comparability with prior work, we follow the evaluation protocols and metrics reported in the corresponding original papers. For Tree-Ring watermarks, we report the detector p-value (i.e., the probability of observing the watermark statistic under the null hypothesis of no watermark) and, for StableSignature, the bit-recovery accuracy. For Gaussian Shading, we report the raw bit accuracy, defined as the fraction of bits in the recovered message that match the target watermark bit string.

\vspace{-5mm}
\subsection{Quantifying Image Quality and Semantic Drift} 
\vspace{-2mm}

To provide a standard quantification of image fidelity, we report perceptual fidelity using standard full-reference image quality metrics of PSNR~\cite{gonzalez2009digital} and SSIM~\cite{wang2004image}.
We must note, however, that these full-reference metrics can be misleading in our setting for attacked watermarked images. In particular, global semantic perturbations intentionally alter image style and, in some cases, content; consequently, PSNR and SSIM may penalize benign semantic changes rather than reflecting meaningful degradation. Prior work often uses CLIP-based scores~\cite{radford2021learning} to assess semantic quality after syntactic perturbations by comparing the generation prompt to the resulting image. In contrast, our goal is to measure the semantic distance between the original watermarked image and its semantically perturbed counterpart. We therefore consider multiple caption-based approaches to better quantify semantic drift.

More specifically, we adopt multimodal text-image measurements. First, we use BLIP~\cite{li2023blip} to generate captions for both the original and attacked images, and compute the cosine similarity between the resulting captions using sentence-transformer embeddings. While this provides a reasonable proxy for semantic preservation, BLIP captions occasionally include spurious or irrelevant metadata, as illustrated in our supplemental material, reflecting biases in its training distribution. We call this score BLIP agreement (BLIPA). Second, we replace BLIP with PaliGemma2~\cite{steiner2024paligemma}, a vision-language model that produces longer, more descriptive captions, and compute caption similarity in the same manner. We call this score VLM agreement (VLMA). Beyond simple captions, we also extract structured semantic information through a scene graph generation process (SGG) \cite{li2024scene} guided by LLMs \cite{kim2024llm4sgg}. We find that graph similarity over such scene graphs largely aligns with cosine similarity on the captions, and thus we defer this metric and its results to the supplemental material.

\vspace{-5mm}
\subsection{Sensitivity to Image-Processing-based Perturbations}
\vspace{-2mm}
\begin{figure}[!t]
  \centering
  \begin{subfigure}[b]{0.32\linewidth}
    \centering
    \includegraphics[width=\linewidth]{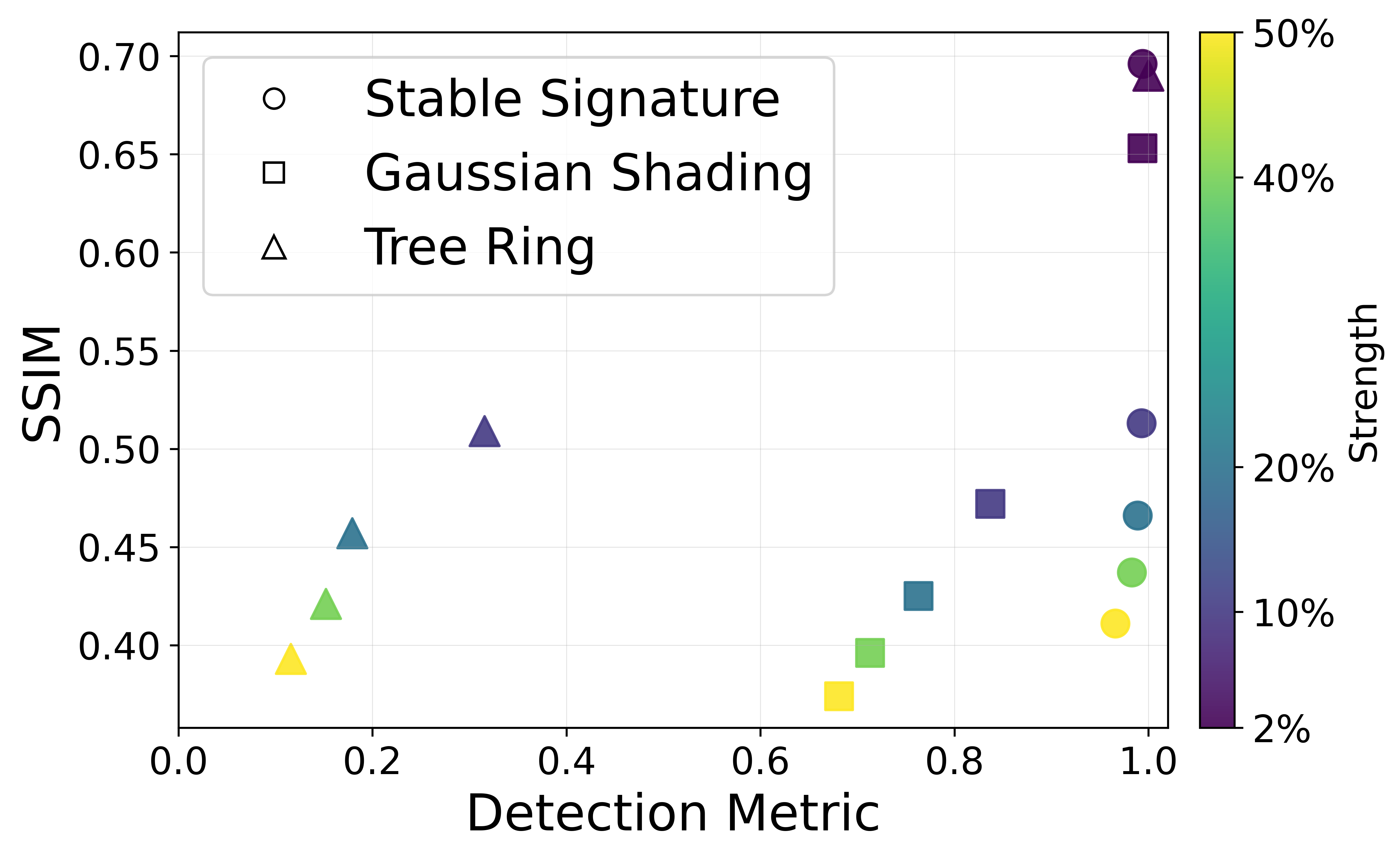}
    \subcaption{Seam Carving}
  \end{subfigure}\hfill
  \begin{subfigure}[b]{0.32\linewidth}
    \centering
    \includegraphics[width=\linewidth]{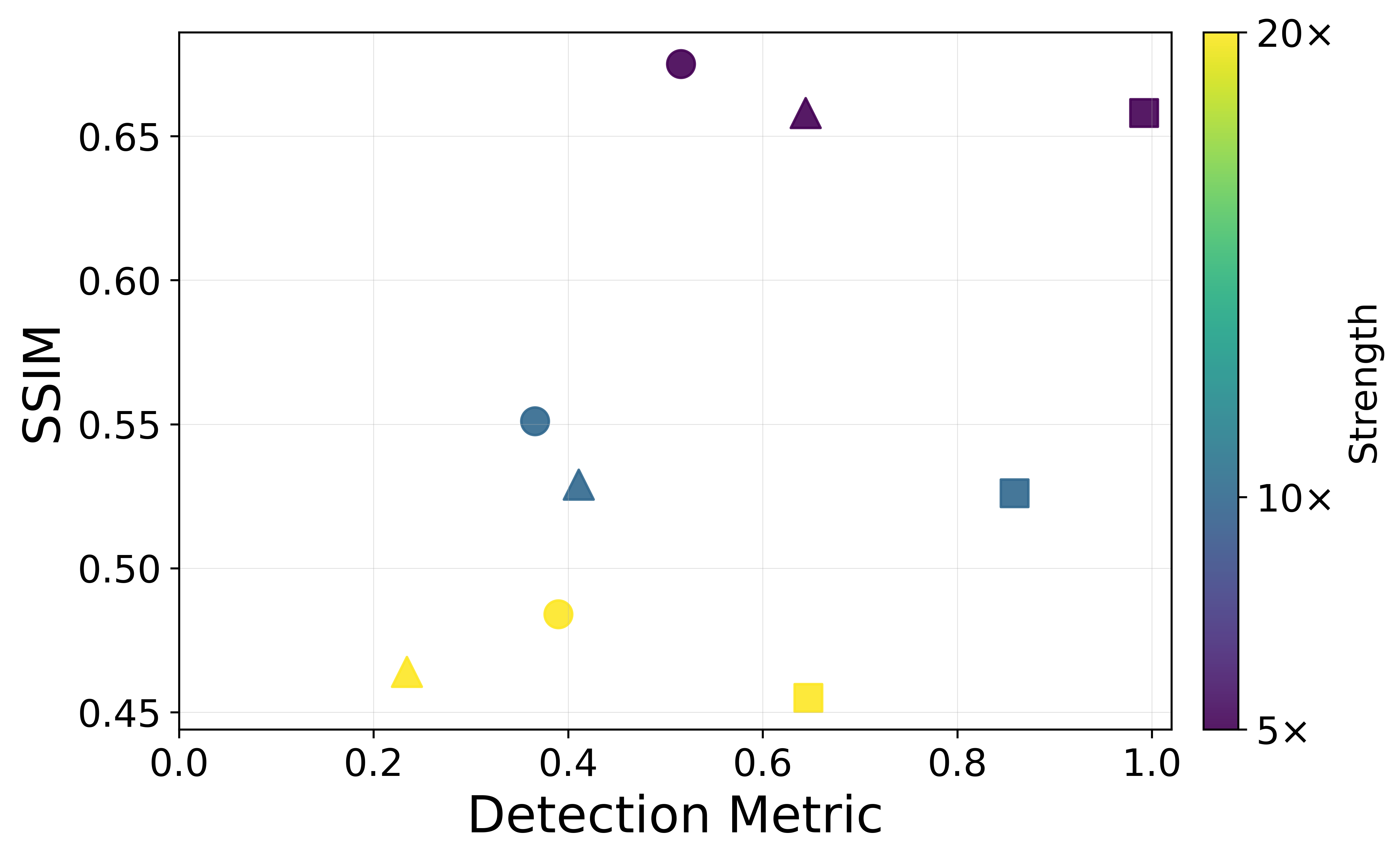}
    \subcaption{Downsampling}
  \end{subfigure}\hfill
  \begin{subfigure}[b]{0.32\linewidth}
    \centering
    \includegraphics[width=\linewidth]{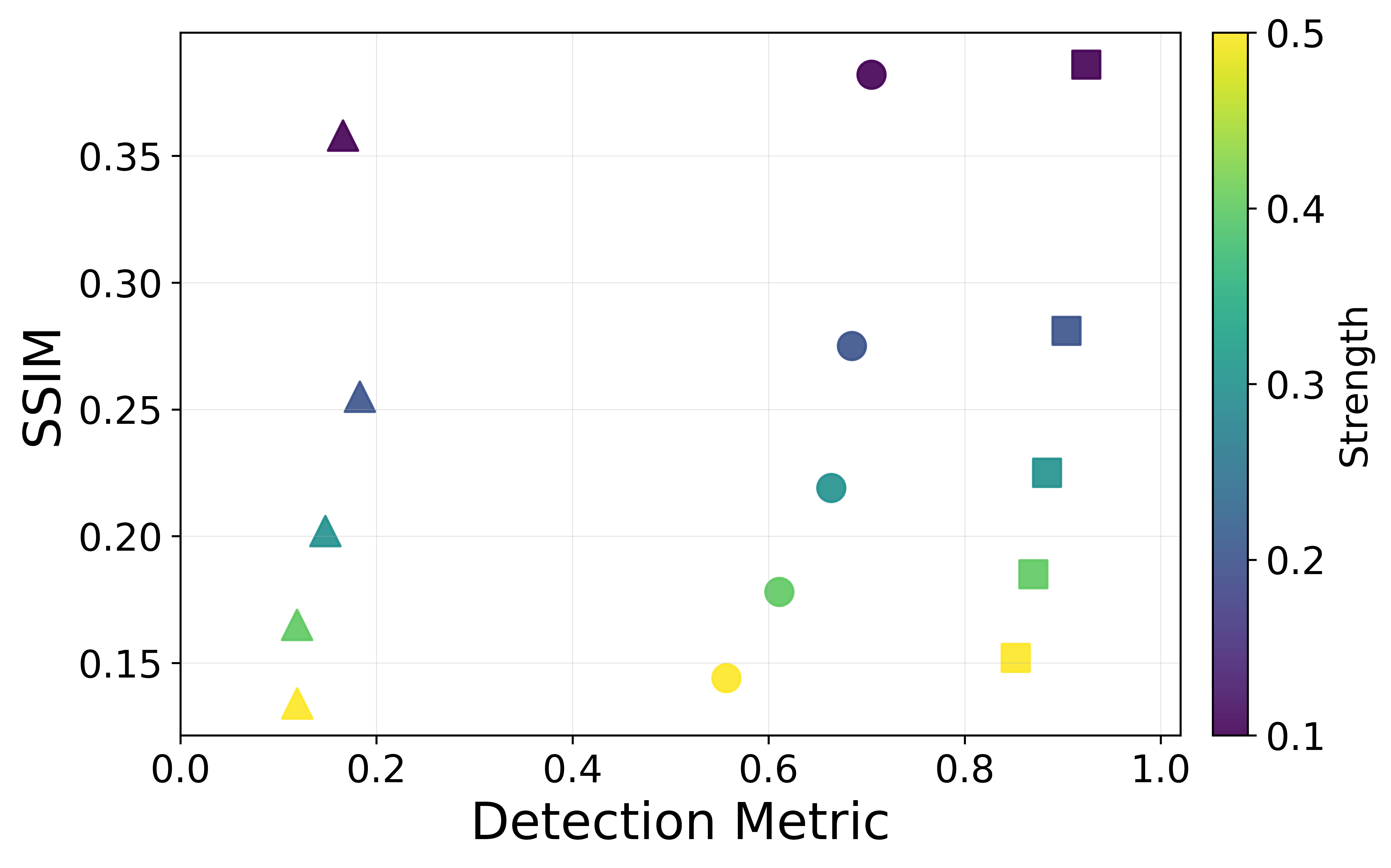}
    \subcaption{Impulse Noise} 
  \end{subfigure}

  \begin{subfigure}[b]{0.32\linewidth}
    \centering
    \includegraphics[width=\linewidth]{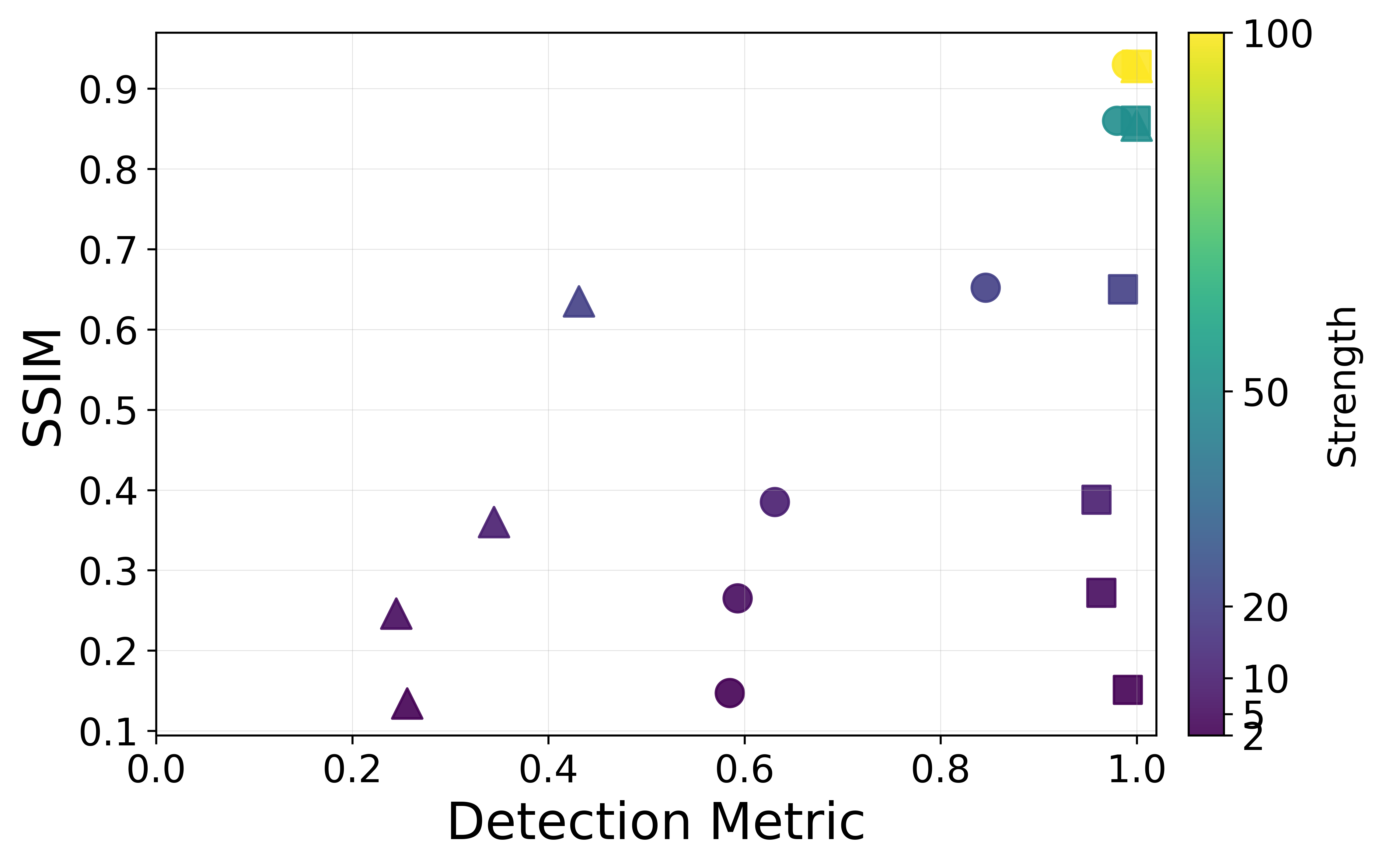}
    \subcaption{Interleaving}
  \end{subfigure}\hfill
  \begin{subfigure}[b]{0.32\linewidth}
    \centering
    \includegraphics[width=\linewidth]{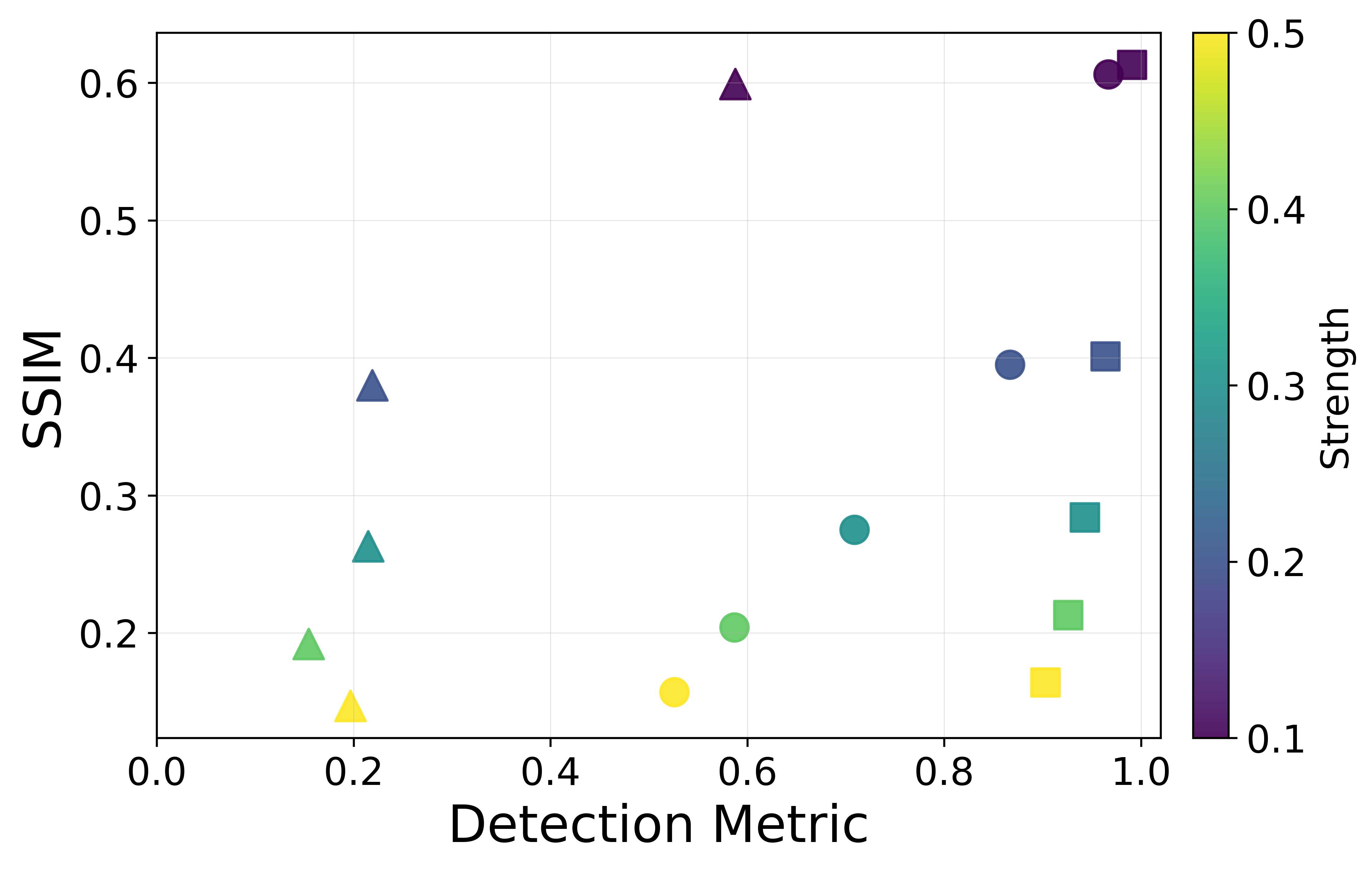}
    \subcaption{Occlusion}
  \end{subfigure}\hfill
  \begin{subfigure}[b]{0.32\linewidth}
    \centering
    \includegraphics[width=\linewidth]{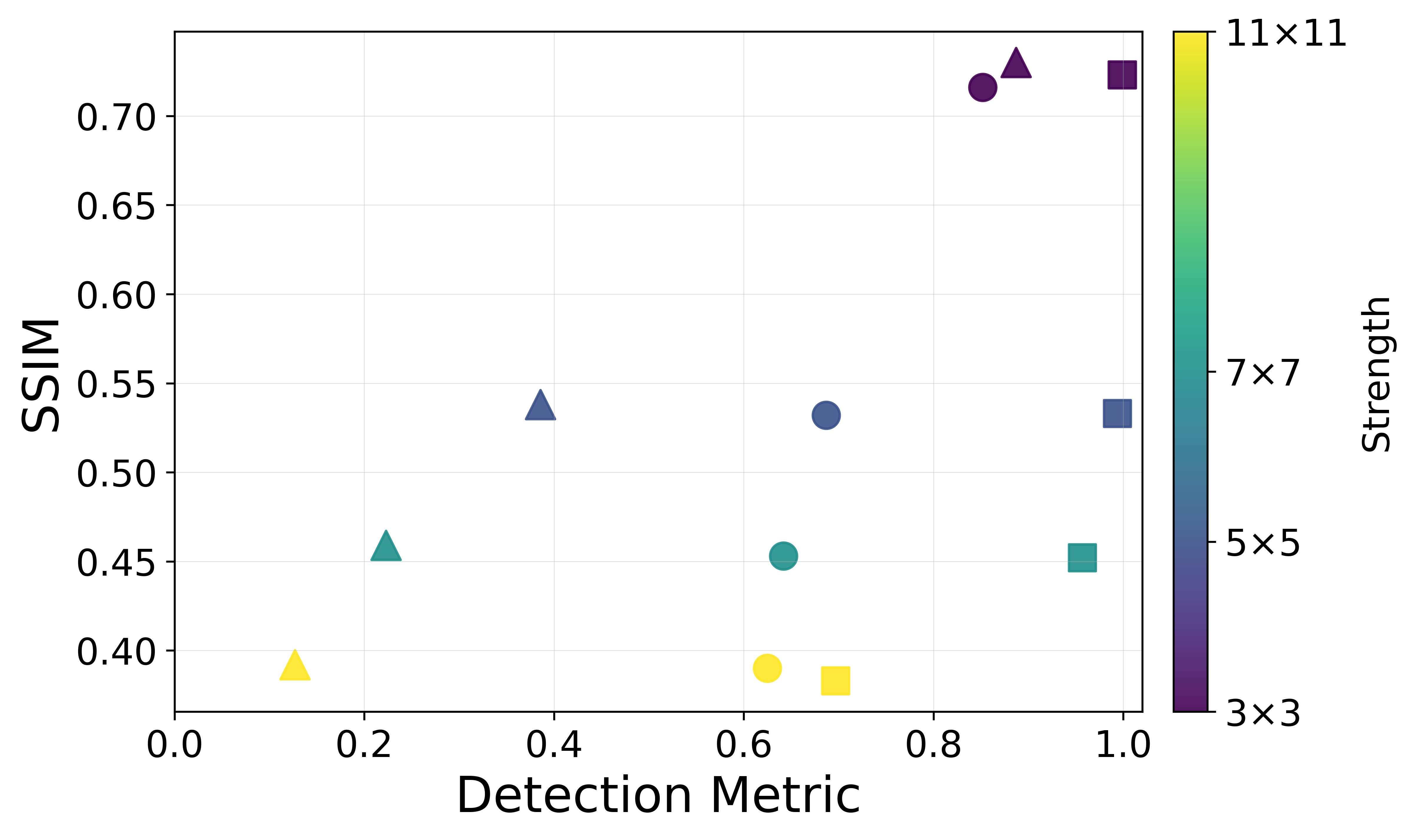}
    \subcaption{Morph. Erosion}
  \end{subfigure}

  \begin{subfigure}[b]{0.32\linewidth}
    \centering
    \includegraphics[width=\linewidth]{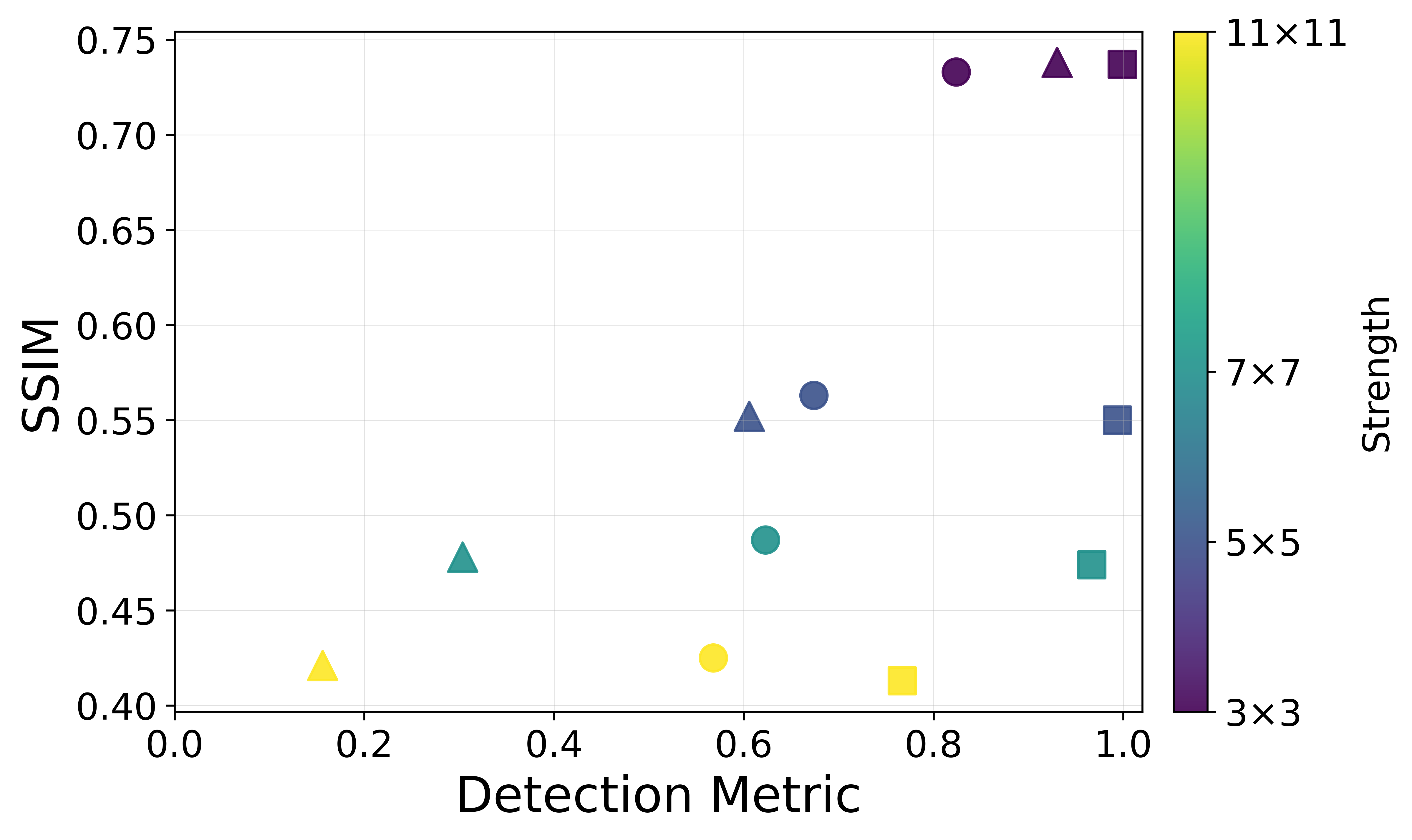}
    \subcaption{Morph. Dilation}
  \end{subfigure}\hfill
  \begin{subfigure}[b]{0.32\linewidth}
    \centering
    \includegraphics[width=\linewidth]{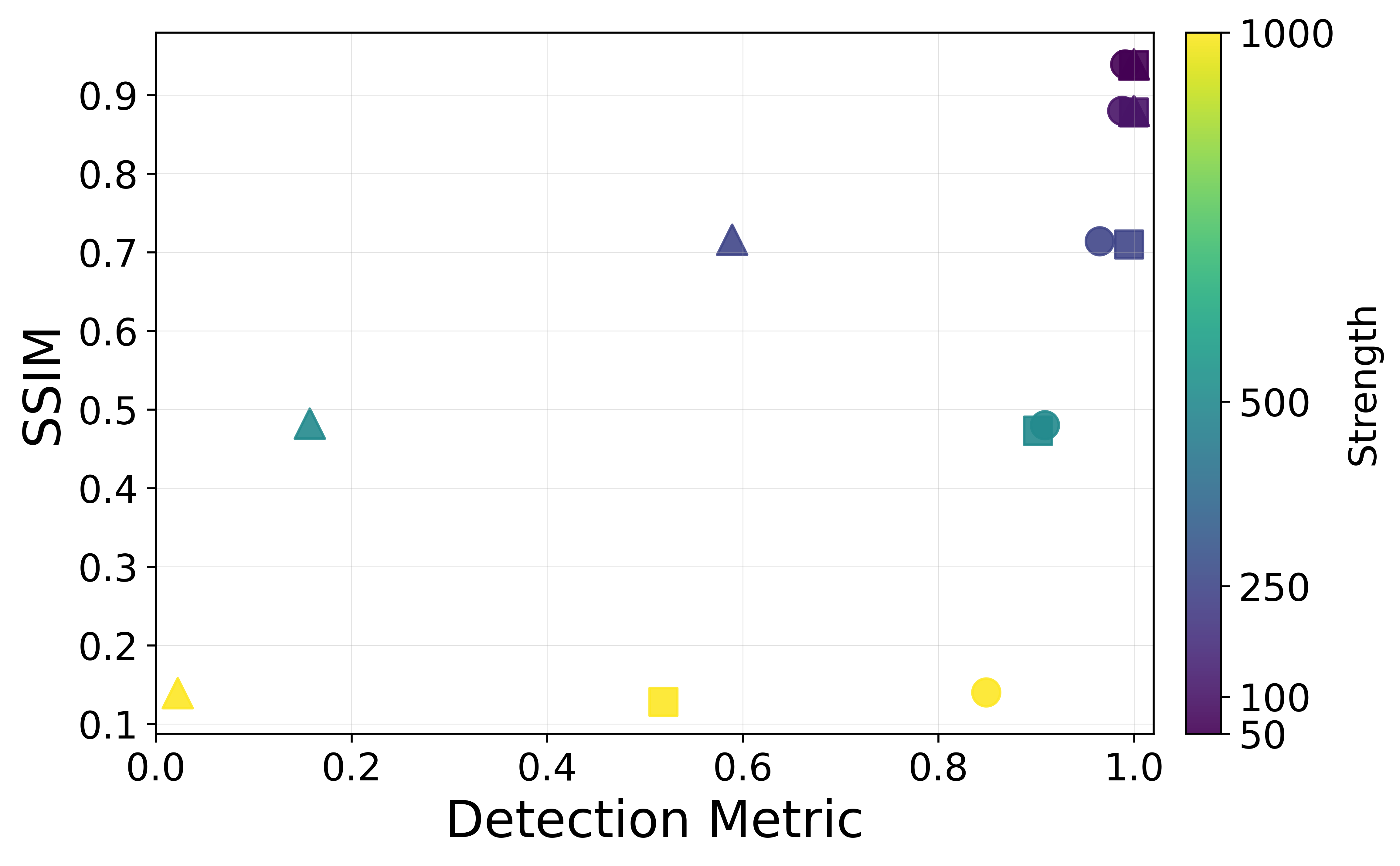}
    \subcaption{Partial Block Shuffling}
  \end{subfigure}\hfill
  \begin{subfigure}[b]{0.32\linewidth}
    \centering
    \includegraphics[width=\linewidth]{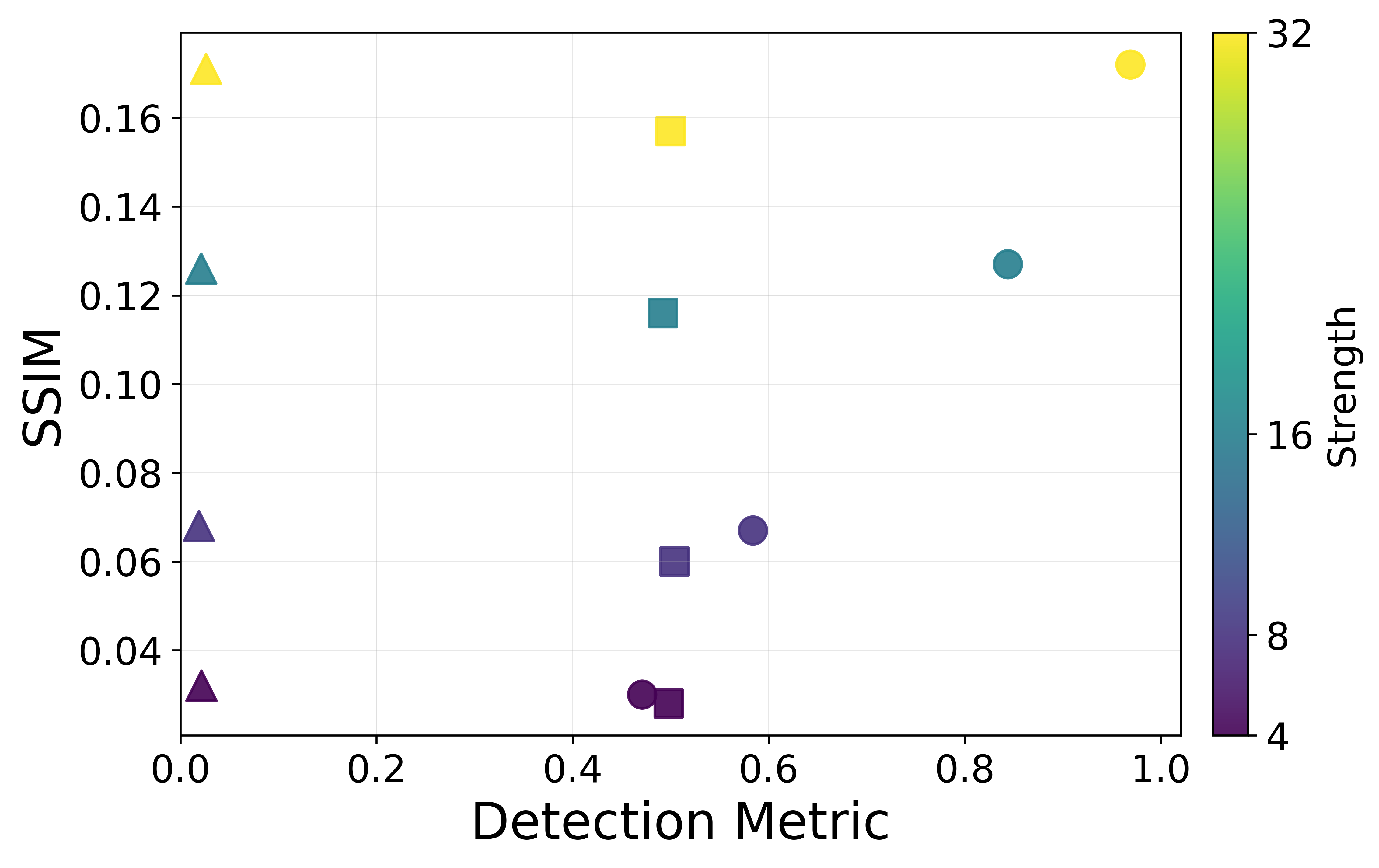}
    \subcaption{Complete Block Shuffling}
  \end{subfigure}
    \vspace{-2mm}
  \caption{Experimental results of watermark detectability vs visual fidelity in SSIM under enhanced image-processing-based perturbations, showing general robustness under these manipulations. Detection metrics are bit accuracy for StableSig/Gaussian shading and the normalized negative logarithm of $p$-value for tree-ring. Color indicates the perturbation strength, and markers indicate the watermarking schemes.  Note that we normalize the $p$-values for understanding the trend of the tree-ring detection metric with perturbation strength. Extended results in tabular form are provided in the supplemental material.}
  \label{fig:3x3grid}
  \vspace{-9mm}
\end{figure}

To establish a strong baseline, we first subject watermarked images to a comprehensive suite of more than ten conventional, method-agnostic perturbations, as elaborated in Sec.~\ref{sec:image_processing}. These perturbations operate at the signal level and include content-aware resizing, downsampling/upsampling, additive and impulse noise, spatial filtering, morphological operations, and block-wise shuffling and have not been previously studied. We illustrate our findings in Fig.~\ref{fig:3x3grid}.

We observe several consistent trends for StableSignature. StableSignature is robust to content-aware resizing, remaining detectable even after removing nearly half the seams, consistent with a watermark that is not localized to specific pixels (potentially aided by resizing augmentations during training). In contrast, detectability degrades rapidly under increasing row occlusion and impulse (shot) noise, with bit accuracy often dropping to 50--60\%. Downsampling--upsampling can eliminate the watermark but at substantial fidelity loss, and morphological operations (erosion/dilation) also significantly reduce detectability with visible degradation. Surprisingly, block shuffling has little effect, suggesting relative insensitivity to block-level spatial permutations. 

For Tree-Ring and Gaussian shading, most conventional perturbations reduce $p$-values for Tree-Ring and bit accuracy for Gaussian shading, but detectability typically remains acceptable. Both methods are sensitive to seam carving: aggressive seam removal can significantly reduce watermark detection metrics, often with noticeable fidelity loss. Localized corruption via black-pixel interleaving has a limited effect overall; row occlusion and impulse (shot) noise only mildly degrade Gaussian shading detectability but substantially hamper Tree-Ring $p$-values and detectability. Morphological operations do not affect Gaussian shading detectability, whereas Tree-Ring detectability reduces when large kernels are used. Unlike StableSignature, both methods are vulnerable to block shuffling, because such permutations violate the generator's spatial statistics and were not accounted for under standard post-processing.
\vspace{-4mm}
\subsection{Sensitivity to Semantic Shifts}
\vspace{-2mm}
\begin{figure}[!t]
    \centering
    \includegraphics[width=\linewidth]{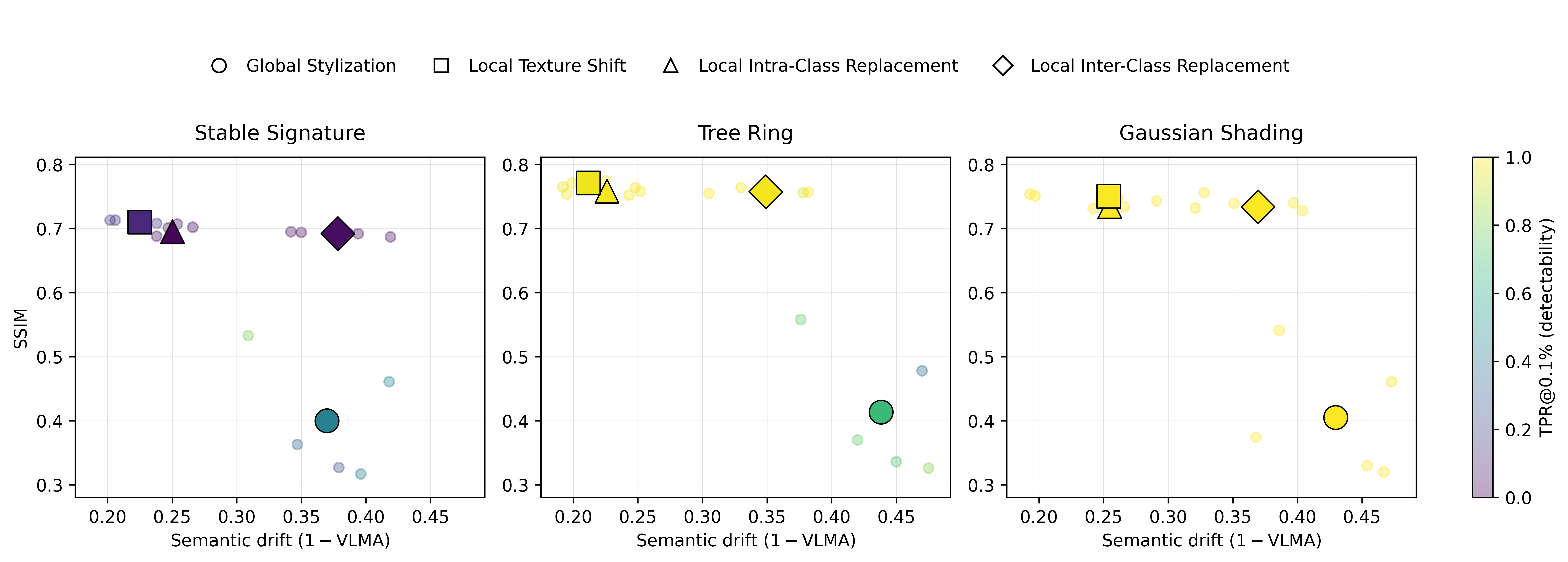}
    \vspace{-5mm}
    \caption{Experimental results of watermark detectability against visual fidelity in SSIM ($y$-axis) and semantic drift ($x$-axis) under semantic perturbations. Color indicates detectability (TPR@0.1\%). Results show watermark detectability can collapse across methods under varying levels of semantic drift, revealing a gap not captured by conventional robustness tests.}
    \label{fig:semantic_tradeoff}
    \vspace{-2mm}
\end{figure}

\begin{table*}[!t]
\centering
\caption{Robustness to semantic manipulation across in-processing watermarking methods (averaged over seeds). Detection metric is bit accuracy for stable signature/Gaussian shading, and the average $p$-value for tree ring. Extended results across multiple seeds for each variant are provided in the supplemental material.}
\vspace{-2mm}
\label{tab:watermark_robustness_avg_stacked}
\resizebox{\textwidth}{!}{
\begin{tabular}{c c c |c c |c c| c c}
\toprule
Perturbation & Variant & Method
& PSNR ($\uparrow$) & SSIM ($\uparrow$) & Detection metric & TPR@0.1\% ($\uparrow$) & VLMA ($\uparrow$) & BLIPA ($\uparrow$) \\
\midrule

\multirow{3}{*}{Global} & \multirow{3}{*}{Stylization}
& Stable Signature~{\small\cite{fernandez2023stable}} & $12.7_{\pm 1.2}$ & $0.40_{\pm 0.05}$ & $57.8_{\pm 6.0}\%$ & $0.44$ & $0.63_{\pm 0.22}$ & $0.56_{\pm 0.10}$ \\
& & Tree Ring~\cite{wen2023tree}        & $12.9_{\pm 1.5}$ & $0.41_{\pm 0.05}$ & $2.7e\!-\!02$ & $0.67$ & $0.56_{\pm 0.24}$ & $0.53_{\pm 0.11}$ \\
& & Gaussian Shading~\cite{yang2024gaussian} & $12.3_{\pm 1.5}$ & $0.40_{\pm 0.05}$ & $98.9_{\pm 0.2}\%$ & $0.99$ & $0.57_{\pm 0.24}$ & $0.55_{\pm 0.11}$ \\
\midrule

\multirow{9}{*}{Local}
& \multirow{3}{*}{Texture Shift}
& Stable Signature~{\small\cite{fernandez2023stable}} & $16.9_{\pm 4.1}$ & $0.71_{\pm 0.10}$ & $50.2_{\pm 7.9}\%$ & $0.10$ & $0.77_{\pm 0.19}$ & $0.78_{\pm 0.11}$ \\
& & Tree Ring~{\small\cite{wen2023tree}}        & $16.6_{\pm 4.5}$ & $0.77_{\pm 0.11}$ & $1.4e\!-\!07$ & $0.97$ & $0.78_{\pm 0.19}$ & $0.76_{\pm 0.13}$ \\
& & Gaussian Shading~{\small\cite{yang2024gaussian}} & $15.8_{\pm 4.5}$ & $0.75_{\pm 0.12}$ & $99.7_{\pm 0.1}\%$ & $1.00$ & $0.74_{\pm 0.25}$ & $0.75_{\pm 0.11}$ \\
\cmidrule(lr){2-9}

& \multirow{3}{*}{Intra-Class}
& Stable Signature~{\small\cite{fernandez2023stable}} & $16.5_{\pm 4.2}$ & $0.69_{\pm 0.10}$ & $49.5_{\pm 7.6}\%$ & $0.01$ & $0.75_{\pm 0.22}$ & $0.80_{\pm 0.09}$ \\
& & Tree Ring~{\small\cite{wen2023tree}}        & $16.0_{\pm 4.4}$ & $0.75_{\pm 0.11}$ & $1.5e\!-\!06$ & $0.98$ & $0.77_{\pm 0.19}$ & $0.76_{\pm 0.14}$ \\
& & Gaussian Shading~{\small\cite{yang2024gaussian}} & $15.5_{\pm 4.3}$ & $0.73_{\pm 0.12}$ & $99.6_{\pm 0.1}\%$ & $1.00$ & $0.74_{\pm 0.25}$ & $0.77_{\pm 0.11}$ \\
\cmidrule(lr){2-9}

& \multirow{3}{*}{Inter-Class}
& Stable Signature~{\small\cite{fernandez2023stable}} & $16.3_{\pm 4.5}$ & $0.69_{\pm 0.11}$ & $49.9_{\pm 7.9}\%$ & $0.03$ & $0.62_{\pm 0.22}$ & $0.67_{\pm 0.16}$ \\
& & Tree Ring~{\small\cite{wen2023tree}}        & $16.0_{\pm 4.6}$ & $0.75_{\pm 0.12}$ & $8.8e\!-\!08$ & $0.98$ & $0.65_{\pm 0.24}$ & $0.65_{\pm 0.18}$ \\
& & Gaussian Shading~{\small\cite{yang2024gaussian}} & $15.0_{\pm 4.3}$ & $0.73_{\pm 0.12}$ & $99.6_{\pm 0.1}\%$ & $1.00$ & $0.63_{\pm 0.25}$ & $0.65_{\pm 0.16}$ \\
\bottomrule
\end{tabular}
}
\label{table:semantic_tradeoff}
\vspace{-5mm}
\end{table*}

Having established robustness under pixel-level perturbations, we next evaluate these schemes under our semantic perturbation framework, which induces progressively larger semantic drift from the original synthesized image, and summarize our results in Fig.~\ref{fig:semantic_tradeoff} and Table~\ref{table:semantic_tradeoff}. Our results show that in-processing watermarking schemes that remain detectable under aggressive signal-level distortions can nevertheless be substantially degraded or rendered ineffective by semantic drift, underscoring the need for robustness evaluations that explicitly model semantically coherent edits rather than only pixel-level corruption.

StableSignature degrades sharply under semantic edits. Across all three local perturbation variants (texture shift, intra-class, and inter-class replacement), bit-recovery accuracy collapses to near chance, remaining tightly concentrated around 49--51\%. Under global semantic perturbations (stylization), performance remains poor, with bit accuracy in the 53--65\% range. Overall, semantic drift renders the decoder-conditioned signature largely non-recoverable. 

Tree-Ring exhibits a distinct failure mode. Under local perturbations (implemented via inpainting within the same model family), Tree-Ring remains statistically detectable, but with substantially weakened evidence: $p$-values increase to the $10^{-10}\sim10^{-6}$ regime across seeds, indicating non-trivial sensitivity despite continued detection. 
This is even true when the local perturbations heavily change the semantics of the scene, reflecting that although tree-ring watermarks change the underlying generating distribution of the LDMs, the semantics might not be entangled with the watermarks in a meaningful way.  In contrast, global semantic perturbations markedly undermine watermark detectability: $p$-values for detection increase further into the $10^{-3}\sim10^{-1}$ range, approaching (and in some settings crossing) typical non-detection thresholds. 

We find that Gaussian shading is the most robust against local and global semantic perturbations among the three in-processing watermarks. It maintains near-perfect message recovery across both local and global semantic perturbations, with bit accuracy consistently high ($\geq$98\% for local and global perturbations). 

\begin{figure}[!t]
    \centering
    \begin{subfigure}[t]{0.33\linewidth}
        \centering
        \includegraphics[width=\linewidth]{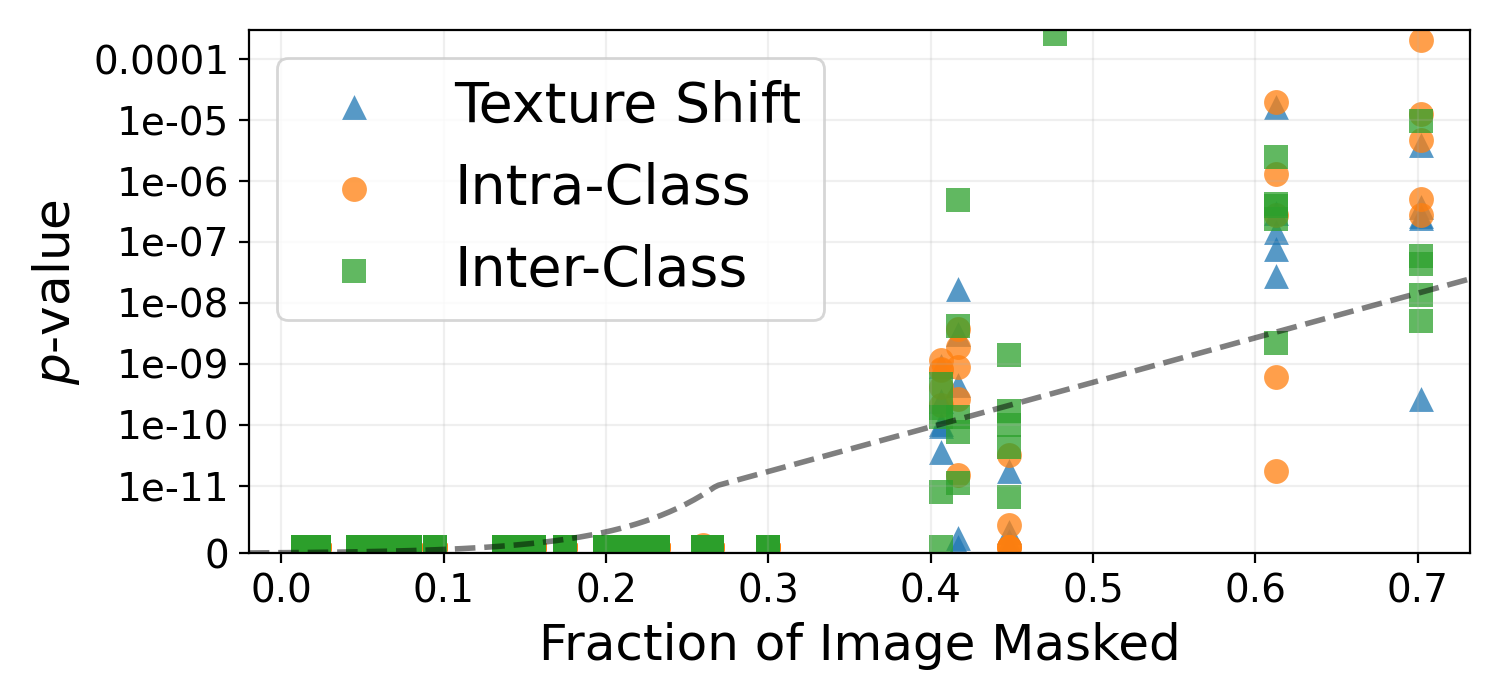}
        \caption{}
    \end{subfigure}\hfill
    \begin{subfigure}[t]{0.33\linewidth}
        \centering
        \includegraphics[width=\linewidth]{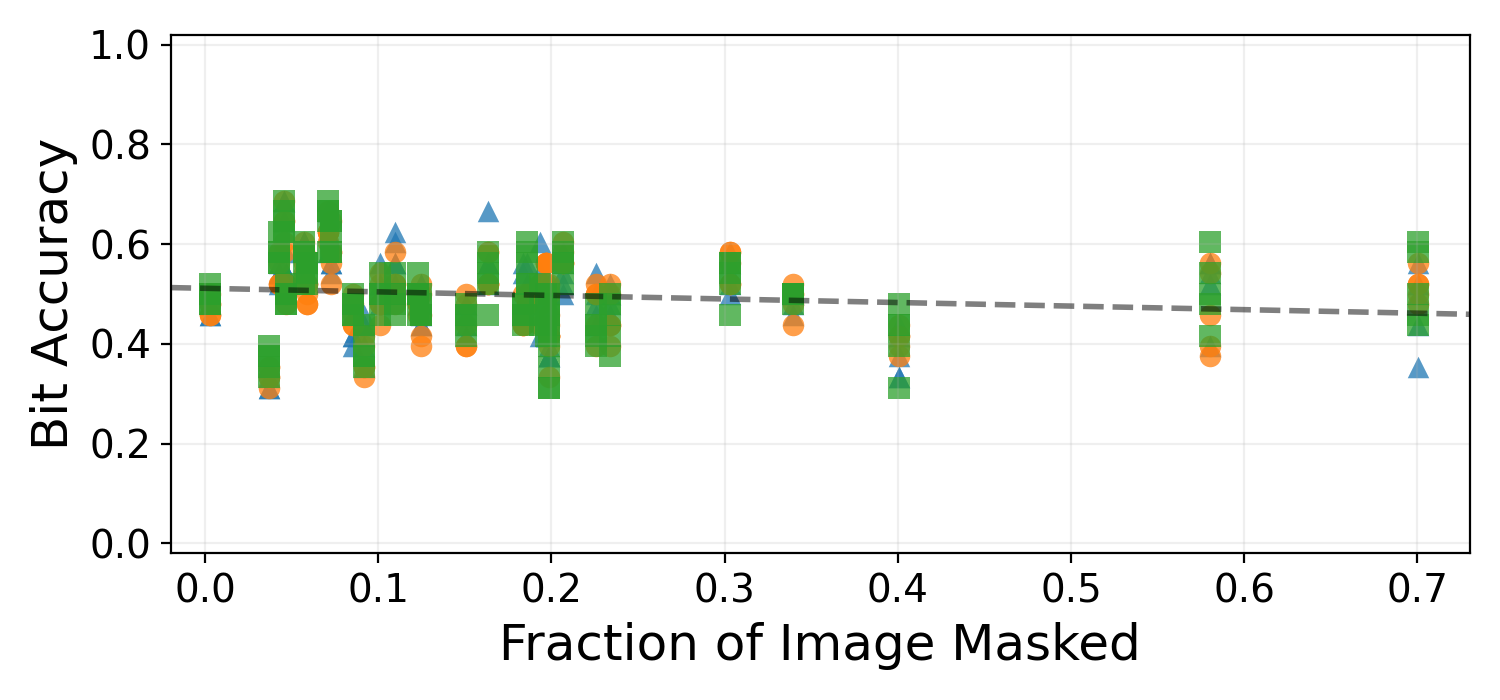}
        \caption{}
    \end{subfigure}
    \begin{subfigure}[t]{0.33\linewidth}
        \centering
        \includegraphics[width=\linewidth]{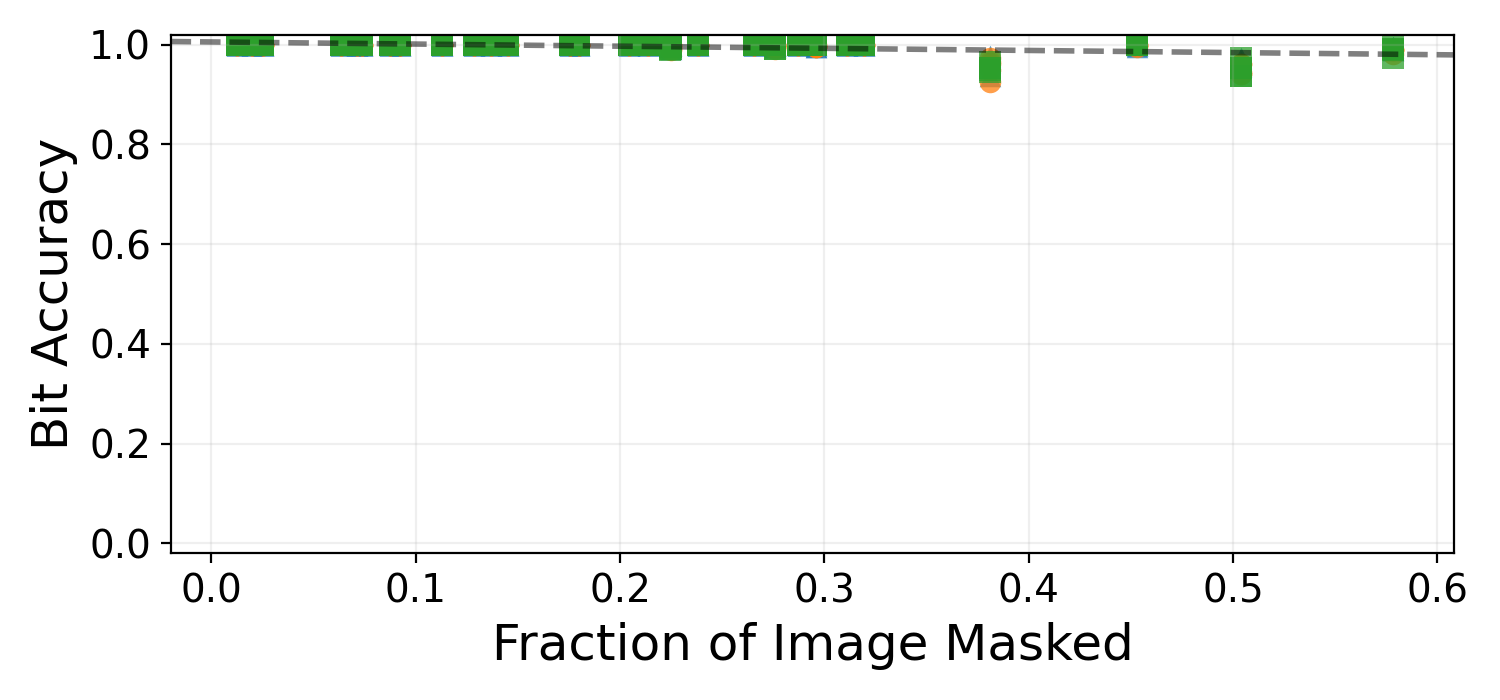}
        \caption{}
    \end{subfigure}
    \vspace{-3mm}
    \caption{Experimental results on watermark detectability under local semantic perturbations as a function of masked area. (a)~$p$-value of the detected tree-ring watermark when an increasing fraction of the image is masked. It is positively correlated with the mask size, indicating reduced detectability as a larger fraction of the image is replaced. (b)~Bit accuracy of the detected stable signature watermark remains near chance across all mask sizes, indicating failure even for small masks. (c)~Bit accuracy of the Gaussian shading watermark remains consistently high, showing minimal sensitivity to mask size.}
    \label{fig:masking_side_by_side}
    \vspace{-7mm}
\end{figure}

\noindent \textbf{Effect of Mask Size.} To study how mask size affects in-processing watermark robustness to semantic shifts, we analyze detection scores as a function of the fraction of pixels replaced by semantic editing. For StableSignature, as illustrated in Fig~\ref{fig:masking_side_by_side}(b), watermark detectability is consistently low across mask sizes, with recovery near chance even for small edited regions. For Tree-Ring, as illustrated in Fig.~\ref{fig:masking_side_by_side}(a), we observe a clear monotonic trend: larger edited regions lead to stronger degradation in detectability. Images dominated by a small number of large foreground objects are particularly susceptible, since editing these objects perturbs a substantial portion of the scene, whereas small, localized edits tend to have a weaker effect in removing the watermark. Conversely, when most of the original image remains intact, and the overall semantics are largely preserved, Tree-Ring often remains detectable. Gaussian Shading Fig.~\ref{fig:masking_side_by_side}(c) exhibits no straightforward dependence on mask size, consistent with its overall robustness to local semantic perturbations. Collectively, these results suggest that, for at least some schemes, the watermark may not have a stable, interpretable entanglement with the semantics of the scene, in contrast to common assumptions in prior work, and robustness against semantic manipulation varies significantly depending on the method. 

\noindent \textbf{Cross-family Local Semantic Edits.} To disentangle robustness effects attributable to the diffusion family used for local semantic perturbations, we evaluate an inpainting model drawn from a different generator family from the watermarked model: Kandinsky 2.1~\cite{razzhigaev2023kandinsky}. This reflects a realistic deployment setting, since users may apply semantic edits using any commercially available, off-the-shelf diffusion model, rather than from the original generator family. Moreover, out-of-family inpainting can constitute a stronger perturbation, as distributional mismatch between the watermarking model and the editing model may induce larger representation shifts and more effectively disrupt watermark recovery, particularly for schemes that rely on the generator’s learned data distribution. As shown in Table~\ref{tab:watermark_robustness_treering_same_vs_cross}, cross-family semantic edits degrade detection statistics more than same-family inpainting, but typically do not eliminate the watermark completely. This suggests that as long as a substantial fraction of pixels in the scene remains drawn from the original generator’s distribution, watermark detectability may remain high.

\begin{table}[!t]
\centering
\caption{Local semantic perturbations using the same or different diffusion model family. Detection metric is bit accuracy for stable signature/Gaussian shading, and the average $p$-value for tree ring. Cross-family perturbations may degrade detection metrics more than same-family perturbations.}
\vspace{-3mm}
\label{tab:watermark_robustness_treering_same_vs_cross}
\resizebox{\linewidth}{!}{
\begin{tabular}{c c c c c c c}
\toprule
\multirow{2}{*}{Attack Category}
& \multicolumn{3}{c}{Same Family}
& \multicolumn{3}{c}{Cross Family} \\
\cmidrule(lr){2-4}\cmidrule(lr){5-7}
& PSNR ($\uparrow$) & SSIM ($\uparrow$) & Detection Metric
& PSNR ($\uparrow$) & SSIM ($\uparrow$) & Detection Metric \\
\midrule

Stable Signature~\cite{fernandez2023stable}.
& $16.29$ & $0.68$ & $50.10\%$
& $15.43$ & $0.64$ & $53.49\%$\\
Tree Ring~\cite{wen2023tree}.
& $15.83$ & $0.75$ & $3.88e-07$
& $15.18$ & $0.63$ & $9.08e-07$ \\
Gaussian Shading~\cite{yang2024gaussian}.
& $14.87$ & $0.72$ & $99.60\%$
& $13.85$ & $0.63$ & $98.43\%$ \\

\bottomrule
\end{tabular}
}
\vspace{-6mm}
\end{table}

\vspace{-4mm}
\section{Discussion and Conclusion}
\vspace{-3mm}
In this paper, we identify an underexplored vulnerability in representative in-processing generative image watermarks. While these methods are robust to a broad suite of syntactic perturbations, they can be substantially degraded by semantic manipulations that alter high-level scene content. This suggests that coupling watermark evidence to a generator’s semantic machinery is a double-edged sword: semantic entanglement can confer resilience to signal-level noise yet increase susceptibility to content-level edits. Our results challenge the prevailing provenance evaluation protocols and motivate benchmarks that explicitly quantify robustness as a function of semantic drift and the corresponding effects on watermark detectability.
Building on this first-step effort, we consider providing a complete mechanistic account of why semantic manipulation affects in-processing methods unevenly as a key future direction. Enabled by our benchmark, researchers may determine which higher-level image variations watermarks exploit in the generative setting. Answering this question may help bridge the gap between traditional signal-processing watermarking theory and modern GenAI watermarking practice. Looking forward, more reliable authentication will likely require multi-layered defenses, including designs that remain detectable under both syntactic and semantic edits.

\bibliographystyle{splncs04}
\bibliography{main_bib}

\clearpage

\appendix
\section{Appendix}

\subsection{Comprehensive Image Processing Perturbation Results}

This section provides the full results of stress-testing the watermark against the complete suite of conventional attacks.

\begin{table*}[!ht]
\centering
\caption{Robustness to improved baseline image-processing-based perturbations}
\label{tab:<label>}
\resizebox{\textwidth}{!}{
\begin{tabular}{c c c c c | c c c | c c c}
\toprule
\multirow{2}{*}{Attack Category}
& \multirow{2}{*}{\centering Strength}
& \multicolumn{3}{c|}{Stable Signature}
& \multicolumn{3}{c|}{Tree Ring}
& \multicolumn{3}{c}{Gaussian Shading}
\\
\cmidrule(lr){3-5}\cmidrule(lr){6-8}\cmidrule(lr){9-11}
&
& PSNR ($\uparrow$) & SSIM ($\uparrow$) & Bit Acc ($\uparrow$)
& PSNR ($\uparrow$) & SSIM ($\uparrow$) & $p$-val ($\downarrow$)
& PSNR ($\uparrow$) & SSIM ($\uparrow$) & Bit Acc ($\uparrow$)
\\
\midrule

\multirow{5}{*}{Seam Carving}
& 2\%  & $21.24_{\pm 3.17}$ & $0.69_{\pm 0.11}$ & $0.99_{\pm 0.02}$ & $20.65_{\pm 3.97}$ & $0.69_{\pm 0.13}$ & $1.10e\text{-}21$ & $19.46_{\pm 2.77}$ & $0.65_{\pm 0.11}$ & $0.99_{\pm 0.01}$ \\
& 10\% & $16.06_{\pm 2.63}$ & $0.51_{\pm 0.13}$ & $0.99_{\pm 0.02}$ & $15.83_{\pm 2.92}$ & $0.50_{\pm 0.13}$ & $1.80e\text{-}05$ & $14.63_{\pm 2.70}$ & $0.47_{\pm 0.13}$ & $0.83_{\pm 0.11}$ \\
& 20\% & $14.76_{\pm 2.46}$ & $0.46_{\pm 0.13}$ & $0.98_{\pm 0.02}$ & $14.56_{\pm 2.66}$ & $0.45_{\pm 0.13}$ & $2.00e\text{-}03$ & $13.53_{\pm 2.54}$ & $0.42_{\pm 0.13}$ & $0.76_{\pm 0.10}$ \\
& 40\% & $14.02_{\pm 2.34}$ & $0.43_{\pm 0.13}$ & $0.98_{\pm 0.03}$ & $13.81_{\pm 2.60}$ & $0.42_{\pm 0.13}$ & $5.20e\text{-}03$ & $12.94_{\pm 2.51}$ & $0.39_{\pm 0.13}$ & $0.71_{\pm 0.10}$ \\
& 50\% & $13.42_{\pm 2.23}$ & $0.41_{\pm 0.13}$ & $0.96_{\pm 0.04}$ & $13.26_{\pm 2.57}$ & $0.39_{\pm 0.12}$ & $1.80e\text{-}02$ & $12.52_{\pm 2.43}$ & $0.37_{\pm 0.13}$ & $0.68_{\pm 0.10}$ \\
\midrule

\multirow{3}{*}{Downsampling}
& $5\times$  & $23.76_{\pm 3.00}$ & $0.67_{\pm 0.10}$ & $0.51_{\pm 0.06}$ & $22.98_{\pm 2.67}$ & $0.65_{\pm 0.10}$ & $2.10e\text{-}10$ & $22.86_{\pm 2.73}$ & $0.65_{\pm 0.11}$ & $0.99_{\pm 0.01}$ \\
& $10\times$ & $20.70_{\pm 2.79}$ & $0.55_{\pm 0.13}$ & $0.36_{\pm 0.04}$ & $20.06_{\pm 2.56}$ & $0.52_{\pm 0.12}$ & $6.80e\text{-}07$ & $19.91_{\pm 2.74}$ & $0.52_{\pm 0.13}$ & $0.85_{\pm 0.05}$ \\
& $20\times$ & $18.23_{\pm 2.40}$ & $0.48_{\pm 0.13}$ & $0.39_{\pm 0.02}$ & $17.76_{\pm 2.52}$ & $0.46_{\pm 0.13}$ & $3.00e\text{-}04$ & $17.51_{\pm 2.60}$ & $0.45_{\pm 0.13}$ & $0.64_{\pm 0.04}$ \\
\midrule

\multirow{5}{*}{Impulse noise}
& 0.1 & $15.75_{\pm 0.77}$ & $0.38_{\pm 0.08}$ & $0.70_{\pm 0.12}$ & $15.19_{\pm 0.99}$ & $0.35_{\pm 0.09}$ & $3.24e\text{-}03$ & $15.21_{\pm 1.08}$ & $0.38_{\pm 0.10}$ & $0.92_{\pm 0.09}$ \\
& 0.2 & $12.74_{\pm 0.77}$ & $0.27_{\pm 0.07}$ & $0.68_{\pm 0.12}$ & $12.19_{\pm 0.98}$ & $0.25_{\pm 0.08}$ & $1.79e\text{-}03$ & $12.19_{\pm 1.08}$ & $0.28_{\pm 0.09}$ & $0.90_{\pm 0.09}$ \\
& 0.3 & $10.98_{\pm 0.77}$ & $0.21_{\pm 0.06}$ & $0.66_{\pm 0.10}$ & $10.42_{\pm 0.98}$ & $0.20_{\pm 0.06}$ & $6.04e\text{-}03$ & $10.43_{\pm 1.07}$ & $0.22_{\pm 0.08}$ & $0.88_{\pm 0.10}$ \\
& 0.4 & $9.73_{\pm 0.77}$ & $0.17_{\pm 0.05}$ & $0.61_{\pm 0.10}$ & $9.17_{\pm 0.98}$ & $0.16_{\pm 0.05}$ & $1.64e\text{-}02$ & $9.18_{\pm 1.07}$ & $0.18_{\pm 0.07}$ & $0.87_{\pm 0.10}$ \\
& 0.5 & $8.76_{\pm 0.77}$ & $0.14_{\pm 0.04}$ & $0.55_{\pm 0.08}$ & $8.20_{\pm 0.98}$ & $0.13_{\pm 0.05}$ & $1.63e\text{-}02$ & $8.21_{\pm 1.07}$ & $0.15_{\pm 0.06}$ & $0.85_{\pm 0.11}$ \\
\midrule

\multirow{5}{*}{Occlusion}
& 0.1 & $15.79_{\pm 0.99}$ & $0.60_{\pm 0.05}$ & $0.96_{\pm 0.06}$ & $15.34_{\pm 1.03}$ & $0.59_{\pm 0.05}$ & $1.52e\text{-}09$ & $15.31_{\pm 1.10}$ & $0.61_{\pm 0.06}$ & $0.99_{\pm 0.01}$ \\
& 0.2 & $12.78_{\pm 0.85}$ & $0.39_{\pm 0.06}$ & $0.86_{\pm 0.10}$ & $12.26_{\pm 0.99}$ & $0.38_{\pm 0.07}$ & $5.18e\text{-}04$ & $12.24_{\pm 1.11}$ & $0.40_{\pm 0.08}$ & $0.96_{\pm 0.04}$ \\
& 0.3 & $11.02_{\pm 0.85}$ & $0.27_{\pm 0.06}$ & $0.70_{\pm 0.10}$ & $10.46_{\pm 0.98}$ & $0.26_{\pm 0.07}$ & $5.99e\text{-}04$ & $10.46_{\pm 1.13}$ & $0.28_{\pm 0.08}$ & $0.94_{\pm 0.05}$ \\
& 0.4 & $9.77_{\pm 0.80}$ & $0.20_{\pm 0.05}$ & $0.58_{\pm 0.08}$ & $9.21_{\pm 0.98}$ & $0.19_{\pm 0.06}$ & $4.83e\text{-}03$ & $9.21_{\pm 1.14}$ & $0.21_{\pm 0.07}$ & $0.92_{\pm 0.06}$ \\
& 0.5 & $8.81_{\pm 0.80}$ & $0.15_{\pm 0.04}$ & $0.52_{\pm 0.07}$ & $8.26_{\pm 0.98}$ & $0.14_{\pm 0.05}$ & $1.11e\text{-}03$ & $8.25_{\pm 1.13}$ & $0.16_{\pm 0.06}$ & $0.90_{\pm 0.07}$ \\
\midrule

\multirow{6}{*}{Interleaving}
& 100 & $25.83_{\pm 0.79}$ & $0.93_{\pm 0.01}$ & $0.99_{\pm 0.02}$ & $25.20_{\pm 0.99}$ & $0.92_{\pm 0.01}$ & $7.05e\text{-}25$ & $25.23_{\pm 1.10}$ & $0.93_{\pm 0.01}$ & $1.00_{\pm 0.00}$ \\
& 50  & $22.76_{\pm 0.78}$ & $0.86_{\pm 0.02}$ & $0.98_{\pm 0.04}$ & $22.18_{\pm 0.98}$ & $0.85_{\pm 0.02}$ & $7.13e\text{-}18$ & $22.19_{\pm 1.07}$ & $0.86_{\pm 0.02}$ & $0.99_{\pm 0.00}$ \\
& 20  & $18.78_{\pm 0.77}$ & $0.65_{\pm 0.05}$ & $0.84_{\pm 0.11}$ & $18.21_{\pm 0.98}$ & $0.63_{\pm 0.06}$ & $3.40e\text{-}07$ & $18.22_{\pm 1.07}$ & $0.65_{\pm 0.06}$ & $0.98_{\pm 0.02}$ \\
& 10  & $15.75_{\pm 0.76}$ & $0.38_{\pm 0.08}$ & $0.63_{\pm 0.10}$ & $15.20_{\pm 0.98}$ & $0.36_{\pm 0.09}$ & $6.79e\text{-}06$ & $15.20_{\pm 1.07}$ & $0.38_{\pm 0.10}$ & $0.95_{\pm 0.05}$ \\
& 5   & $12.73_{\pm 0.76}$ & $0.26_{\pm 0.07}$ & $0.59_{\pm 0.07}$ & $12.18_{\pm 0.98}$ & $0.24_{\pm 0.08}$ & $2.12e\text{-}04$ & $12.19_{\pm 1.07}$ & $0.27_{\pm 0.09}$ & $0.96_{\pm 0.04}$ \\
& 2   & $8.76_{\pm 0.77}$  & $0.14_{\pm 0.04}$ & $0.58_{\pm 0.14}$ & $8.20_{\pm 0.98}$  & $0.13_{\pm 0.05}$ & $1.44e\text{-}04$ & $8.21_{\pm 1.07}$  & $0.15_{\pm 0.06}$ & $0.99_{\pm 0.01}$ \\
\midrule

\multirow{4}{*}{Morph.\ Erosion}
& $3\times3$   & $20.69_{\pm 2.76}$ & $0.71_{\pm 0.08}$ & $0.85_{\pm 0.07}$ & $20.10_{\pm 2.55}$ & $0.73_{\pm 0.07}$ & $4.92e\text{-}14$ & $20.04_{\pm 2.56}$ & $0.72_{\pm 0.09}$ & $0.99_{\pm 0.00}$ \\
& $5\times5$   & $16.91_{\pm 2.59}$ & $0.53_{\pm 0.12}$ & $0.68_{\pm 0.06}$ & $16.33_{\pm 2.48}$ & $0.53_{\pm 0.12}$ & $1.63e\text{-}06$ & $16.26_{\pm 2.49}$ & $0.53_{\pm 0.13}$ & $0.99_{\pm 0.01}$ \\
& $7\times7$   & $15.03_{\pm 2.42}$ & $0.45_{\pm 0.13}$ & $0.64_{\pm 0.06}$ & $14.49_{\pm 2.41}$ & $0.45_{\pm 0.13}$ & $4.52e\text{-}04$ & $14.41_{\pm 2.42}$ & $0.45_{\pm 0.14}$ & $0.95_{\pm 0.04}$ \\
& $11\times11$ & $12.99_{\pm 2.17}$ & $0.39_{\pm 0.14}$ & $0.62_{\pm 0.06}$ & $12.44_{\pm 2.28}$ & $0.39_{\pm 0.14}$ & $1.24e\text{-}02$ & $12.37_{\pm 2.29}$ & $0.38_{\pm 0.14}$ & $0.69_{\pm 0.05}$ \\
\midrule

\multirow{4}{*}{Morph.\ Dilation}
& $3\times3$   & $20.67_{\pm 2.84}$ & $0.73_{\pm 0.07}$ & $0.82_{\pm 0.07}$ & $20.06_{\pm 2.60}$ & $0.73_{\pm 0.07}$ & $1.10e\text{-}14$ & $19.96_{\pm 2.63}$ & $0.73_{\pm 0.08}$ & $0.99_{\pm 0.00}$ \\
& $5\times5$   & $16.89_{\pm 2.76}$ & $0.56_{\pm 0.11}$ & $0.67_{\pm 0.07}$ & $16.29_{\pm 2.58}$ & $0.55_{\pm 0.11}$ & $8.17e\text{-}10$ & $16.13_{\pm 2.66}$ & $0.55_{\pm 0.12}$ & $0.99_{\pm 0.01}$ \\
& $7\times7$   & $14.97_{\pm 2.66}$ & $0.48_{\pm 0.13}$ & $0.62_{\pm 0.07}$ & $14.47_{\pm 2.56}$ & $0.47_{\pm 0.13}$ & $2.77e\text{-}05$ & $14.26_{\pm 2.64}$ & $0.47_{\pm 0.14}$ & $0.96_{\pm 0.03}$ \\
& $11\times11$ & $12.82_{\pm 2.50}$ & $0.42_{\pm 0.13}$ & $0.56_{\pm 0.06}$ & $12.44_{\pm 2.49}$ & $0.42_{\pm 0.14}$ & $4.55e\text{-}03$ & $12.18_{\pm 2.58}$ & $0.41_{\pm 0.14}$ & $0.76_{\pm 0.06}$ \\
\midrule

\multirow{5}{*}{Partial Block Shuffling}
& 50   & $22.29_{\pm 1.73}$ & $0.93_{\pm 0.00}$ & $0.99_{\pm 0.02}$ & $22.22_{\pm 2.07}$ & $0.93_{\pm 0.00}$ & $9.15e\text{-}22$ & $21.67_{\pm 2.04}$ & $0.93_{\pm 0.00}$ & $1.00_{\pm 0.00}$ \\
& 100  & $19.20_{\pm 1.71}$ & $0.88_{\pm 0.00}$ & $0.98_{\pm 0.03}$ & $19.24_{\pm 1.92}$ & $0.88_{\pm 0.00}$ & $2.15e\text{-}19$ & $18.64_{\pm 2.15}$ & $0.87_{\pm 0.01}$ & $1.00_{\pm 0.00}$ \\
& 250  & $15.13_{\pm 1.61}$ & $0.71_{\pm 0.01}$ & $0.96_{\pm 0.05}$ & $15.28_{\pm 1.87}$ & $0.71_{\pm 0.01}$ & $1.45e\text{-}09$ & $14.60_{\pm 2.10}$ & $0.71_{\pm 0.02}$ & $0.99_{\pm 0.00}$ \\
& 500  & $12.06_{\pm 1.59}$ & $0.48_{\pm 0.02}$ & $0.90_{\pm 0.09}$ & $12.23_{\pm 1.82}$ & $0.48_{\pm 0.03}$ & $4.35e\text{-}03$ & $11.58_{\pm 2.10}$ & $0.47_{\pm 0.03}$ & $0.90_{\pm 0.05}$ \\
& 1000 & $9.09_{\pm 1.58}$  & $0.14_{\pm 0.05}$ & $0.84_{\pm 0.10}$ & $9.20_{\pm 1.84}$  & $0.13_{\pm 0.05}$ & $4.61e\text{-}01$ & $8.57_{\pm 2.10}$  & $0.12_{\pm 0.06}$ & $0.51_{\pm 0.03}$ \\
\midrule

\multirow{4}{*}{Complete Block Shuffling}
& 32 & $9.05_{\pm 1.61}$ & $0.17_{\pm 0.08}$ & $0.96_{\pm 0.05}$ & $9.18_{\pm 1.85}$ & $0.17_{\pm 0.08}$ & $4.04e\text{-}01$ & $8.52_{\pm 2.10}$ & $0.15_{\pm 0.08}$ & $0.50_{\pm 0.02}$ \\
& 16 & $9.00_{\pm 1.59}$ & $0.12_{\pm 0.05}$ & $0.84_{\pm 0.10}$ & $9.12_{\pm 1.84}$ & $0.12_{\pm 0.06}$ & $4.80e\text{-}01$ & $8.46_{\pm 2.08}$ & $0.11_{\pm 0.06}$ & $0.49_{\pm 0.02}$ \\
& 8  & $8.98_{\pm 1.60}$ & $0.06_{\pm 0.03}$ & $0.58_{\pm 0.06}$ & $9.11_{\pm 1.85}$ & $0.06_{\pm 0.03}$ & $5.20e\text{-}01$ & $8.47_{\pm 2.10}$ & $0.06_{\pm 0.03}$ & $0.50_{\pm 0.02}$ \\
& 4  & $8.98_{\pm 1.60}$ & $0.03_{\pm 0.01}$ & $0.47_{\pm 0.04}$ & $9.11_{\pm 1.85}$ & $0.03_{\pm 0.01}$ & $4.75e\text{-}01$ & $8.47_{\pm 2.08}$ & $0.02_{\pm 0.02}$ & $0.49_{\pm 0.02}$ \\
\bottomrule
\end{tabular}
}
\end{table*}

\subsection{Comprehensive Semantic Perturbation Results}

This section provides the full results of stress-testing the watermark against the complete suite of semantic perturbations on a diverse set of images from the MS COCO dataset.

\begin{sidewaystable}[!ht]
\centering
\caption{Robustness to semantic manipulation across in-processing watermarking methods}
\label{tab:watermark_robustness}
\resizebox{\textwidth}{!}{
\begin{tabular}{c c c c c c c c c| c c c c c c| c c c c c c}
\toprule
\multirow{3}{*}{Attack Category}
& \multirow{3}{*}{\centering Variant}
& \multirow{3}{*}{\centering Seed}
& \multicolumn{6}{c|}{Stable Signature}
& \multicolumn{6}{c|}{Tree Ring}
& \multicolumn{6}{c}{Gaussian Shading}
\\
\cmidrule(lr){4-9}\cmidrule(lr){10-15}\cmidrule(lr){16-21}
&
&
& PSNR & SSIM  & Bit Acc  & TPR  & VLMA  & BLIPA
& PSNR  & SSIM & $p$-val & TPR  & VLMA  & BLIPA
& PSNR  & SSIM  & Bit Acc & TPR & VLMA  & BLIPA
\\
&
&
& ($\uparrow$) & ($\uparrow$) & ($\uparrow$) & @0.1\% ($\uparrow$) & ($\uparrow$) & ($\uparrow$)
& ($\uparrow$) &($\uparrow$) & ($\downarrow$) & @0.1\% ($\uparrow$) & ($\uparrow$) & ($\uparrow$)
& ($\uparrow$) & ($\uparrow$) & ($\uparrow$) & @0.1\% ($\uparrow$) & ($\uparrow$) & ($\uparrow$)
\\
\midrule

\multirow{5}{*}{Global}
& \multirow{5}{*}{Stylization}
& 1
  & $12.49_{\pm 1.18}$ & $0.31_{\pm 0.06}$ & $0.57_{\pm 0.05}$ & $0.41$ & $0.60_{\pm 0.21}$ & $0.52_{\pm 0.10}$
  & $12.41_{\pm 1.41}$ & $0.32_{\pm 0.06}$ & $1.54e\text{-}03$ & $0.80$ & $0.52_{\pm 0.23}$ & $0.49_{\pm 0.14}$
  & $11.82_{\pm 1.34}$ & $0.32_{\pm 0.06}$ & $0.99_{\pm 0.01}$ & $1.00$ & $0.53_{\pm 0.24}$ & $0.52_{\pm 0.12}$ \\
&  & 2
  & $12.24_{\pm 1.17}$ & $0.46_{\pm 0.04}$ & $0.58_{\pm 0.05}$ & $0.45$ & $0.58_{\pm 0.22}$ & $0.55_{\pm 0.11}$
  & $12.71_{\pm 1.37}$ & $0.47_{\pm 0.05}$ & $1.06e\text{-}01$ & $0.35$ & $0.53_{\pm 0.24}$ & $0.53_{\pm 0.11}$
  & $12.10_{\pm 1.48}$ & $0.46_{\pm 0.05}$ & $0.98_{\pm 0.01}$ & $0.98$ & $0.52_{\pm 0.25}$ & $0.53_{\pm 0.12}$ \\
&  & 3
  & $13.41_{\pm 1.43}$ & $0.53_{\pm 0.05}$ & $0.65_{\pm 0.07}$ & $0.80$ & $0.69_{\pm 0.21}$ & $0.62_{\pm 0.10}$
  & $13.94_{\pm 1.66}$ & $0.55_{\pm 0.05}$ & $1.51e\text{-}03$ & $0.75$ & $0.62_{\pm 0.25}$ & $0.60_{\pm 0.10}$
  & $13.17_{\pm 1.70}$ & $0.54_{\pm 0.06}$ & $0.99_{\pm 0.01}$ & $1.00$ & $0.61_{\pm 0.26}$ & $0.61_{\pm 0.10}$ \\
&  & 4
  & $12.47_{\pm 1.11}$ & $0.32_{\pm 0.04}$ & $0.53_{\pm 0.05}$ & $0.24$ & $0.62_{\pm 0.21}$ & $0.52_{\pm 0.10}$
  & $12.65_{\pm 1.41}$ & $0.33_{\pm 0.04}$ & $2.48e\text{-}02$ & $0.71$ & $0.55_{\pm 0.24}$ & $0.50_{\pm 0.11}$
  & $12.14_{\pm 1.44}$ & $0.33_{\pm 0.04}$ & $0.98_{\pm 0.02}$ & $1.00$ & $0.54_{\pm 0.24}$ & $0.51_{\pm 0.11}$ \\
&  & 5
  & $13.04_{\pm 1.39}$ & $0.36_{\pm 0.06}$ & $0.54_{\pm 0.05}$ & $0.30$ & $0.65_{\pm 0.22}$ & $0.60_{\pm 0.10}$
  & $12.82_{\pm 1.64}$ & $0.37_{\pm 0.06}$ & $1.13e\text{-}03$ & $0.76$ & $0.58_{\pm 0.24}$ & $0.56_{\pm 0.11}$
  & $12.40_{\pm 1.61}$ & $0.37_{\pm 0.05}$ & $0.99_{\pm 0.01}$ & $1.00$ & $0.63_{\pm 0.22}$ & $0.57_{\pm 0.09}$ \\
\midrule

\multirow{15}{*}{Local}

& \multirow{5}{*}{Texture Shift}  & 1
& $16.99_{\pm 4.14}$ & $0.70_{\pm 0.10}$ & $0.48_{\pm 0.07}$ & $0.12$ & $0.76_{\pm 0.20}$ & $0.78_{\pm 0.11}$
& $16.47_{\pm 4.54}$ & $0.77_{\pm 0.12}$ & $6.54e\text{-}07$ & $0.90$ & $0.79_{\pm 0.20}$ & $0.76_{\pm 0.14}$
& $16.04_{\pm 4.42}$ & $0.74_{\pm 0.12}$ & $0.99_{\pm 0.00}$ & $1.00$ & $0.73_{\pm 0.26}$ & $0.73_{\pm 0.10}$ \\
&                                 & 2
& $17.21_{\pm 4.37}$ & $0.71_{\pm 0.11}$ & $0.49_{\pm 0.07}$ & $0.03$ & $0.77_{\pm 0.22}$ & $0.80_{\pm 0.11}$
& $17.18_{\pm 4.42}$ & $0.77_{\pm 0.12}$ & $1.28e\text{-}08$ & $0.99$ & $0.77_{\pm 0.20}$ & $0.77_{\pm 0.13}$
& $16.33_{\pm 4.42}$ & $0.75_{\pm 0.12}$ & $0.99_{\pm 0.01}$ & $1.00$ & $0.80_{\pm 0.22}$ & $0.75_{\pm 0.12}$ \\
&                                 & 3
& $16.45_{\pm 4.22}$ & $0.70_{\pm 0.10}$ & $0.51_{\pm 0.07}$ & $0.06$ & $0.74_{\pm 0.22}$ & $0.74_{\pm 0.13}$
& $16.29_{\pm 4.73}$ & $0.76_{\pm 0.11}$ & $4.82e\text{-}09$ & $1.00$ & $0.78_{\pm 0.16}$ & $0.73_{\pm 0.13}$
& $15.29_{\pm 4.72}$ & $0.74_{\pm 0.12}$ & $0.99_{\pm 0.01}$ & $1.00$ & $0.70_{\pm 0.27}$ & $0.72_{\pm 0.11}$ \\
&                                 & 4
& $17.18_{\pm 4.15}$ & $0.71_{\pm 0.10}$ & $0.51_{\pm 0.08}$ & $0.16$ & $0.79_{\pm 0.18}$ & $0.80_{\pm 0.09}$
& $16.53_{\pm 4.71}$ & $0.77_{\pm 0.11}$ & $1.76e\text{-}08$ & $0.98$ & $0.78_{\pm 0.21}$ & $0.79_{\pm 0.14}$
& $15.80_{\pm 4.70}$ & $0.75_{\pm 0.11}$ & $0.99_{\pm 0.00}$ & $1.00$ & $0.67_{\pm 0.30}$ & $0.79_{\pm 0.10}$ \\
&                                 & 5
& $17.01_{\pm 4.06}$ & $0.71_{\pm 0.10}$ & $0.51_{\pm 0.08}$ & $0.16$ & $0.79_{\pm 0.15}$ & $0.79_{\pm 0.12}$
& $16.56_{\pm 4.56}$ & $0.77_{\pm 0.11}$ & $1.12e\text{-}08$ & $0.98$ & $0.80_{\pm 0.16}$ & $0.76_{\pm 0.13}$
& $16.01_{\pm 4.49}$ & $0.75_{\pm 0.12}$ & $0.99_{\pm 0.00}$ & $1.00$ & $0.80_{\pm 0.21}$ & $0.77_{\pm 0.12}$ \\
\cmidrule(lr){2-21}

& \multirow{5}{*}{Intra-Class}    & 1
& $16.21_{\pm 4.33}$ & $0.68_{\pm 0.11}$ & $0.49_{\pm 0.08}$ & $0.03$ & $0.76_{\pm 0.20}$ & $0.79_{\pm 0.09}$
& $15.68_{\pm 4.37}$ & $0.75_{\pm 0.12}$ & $5.77e\text{-}08$ & $0.98$ & $0.80_{\pm 0.17}$ & $0.76_{\pm 0.15}$
& $15.37_{\pm 4.40}$ & $0.73_{\pm 0.12}$ & $0.99_{\pm 0.01}$ & $1.00$ & $0.75_{\pm 0.24}$ & $0.79_{\pm 0.10}$ \\
&                                 & 2
& $16.58_{\pm 4.28}$ & $0.69_{\pm 0.10}$ & $0.48_{\pm 0.07}$ & $0.00$ & $0.75_{\pm 0.22}$ & $0.79_{\pm 0.11}$
& $16.23_{\pm 4.35}$ & $0.75_{\pm 0.11}$ & $1.51e\text{-}07$ & $0.99$ & $0.74_{\pm 0.21}$ & $0.77_{\pm 0.13}$
& $15.66_{\pm 4.32}$ & $0.73_{\pm 0.12}$ & $0.99_{\pm 0.01}$ & $1.00$ & $0.73_{\pm 0.26}$ & $0.78_{\pm 0.12}$ \\
&                                 & 3
& $16.37_{\pm 4.34}$ & $0.69_{\pm 0.11}$ & $0.49_{\pm 0.07}$ & $0.00$ & $0.75_{\pm 0.22}$ & $0.78_{\pm 0.09}$
& $15.99_{\pm 4.44}$ & $0.75_{\pm 0.12}$ & $1.04e\text{-}06$ & $0.98$ & $0.75_{\pm 0.21}$ & $0.75_{\pm 0.15}$
& $15.54_{\pm 4.35}$ & $0.72_{\pm 0.12}$ & $0.99_{\pm 0.01}$ & $1.00$ & $0.74_{\pm 0.26}$ & $0.74_{\pm 0.14}$ \\
&                                 & 4
& $16.80_{\pm 4.16}$ & $0.70_{\pm 0.10}$ & $0.50_{\pm 0.07}$ & $0.00$ & $0.73_{\pm 0.25}$ & $0.81_{\pm 0.10}$
& $16.19_{\pm 4.46}$ & $0.76_{\pm 0.11}$ & $8.99e\text{-}09$ & $1.00$ & $0.75_{\pm 0.22}$ & $0.76_{\pm 0.15}$
& $15.39_{\pm 4.45}$ & $0.73_{\pm 0.12}$ & $0.99_{\pm 0.01}$ & $1.00$ & $0.75_{\pm 0.25}$ & $0.78_{\pm 0.10}$ \\
&                                 & 5
& $16.86_{\pm 4.17}$ & $0.70_{\pm 0.10}$ & $0.50_{\pm 0.07}$ & $0.03$ & $0.75_{\pm 0.23}$ & $0.80_{\pm 0.08}$
& $16.02_{\pm 4.46}$ & $0.76_{\pm 0.12}$ & $6.64e\text{-}06$ & $0.97$ & $0.80_{\pm 0.16}$ & $0.76_{\pm 0.14}$
& $15.64_{\pm 4.21}$ & $0.74_{\pm 0.12}$ & $0.99_{\pm 0.00}$ & $1.00$ & $0.74_{\pm 0.26}$ & $0.79_{\pm 0.12}$ \\
\cmidrule(lr){2-21}

& \multirow{5}{*}{Inter-Class}    & 1
& $16.29_{\pm 4.62}$ & $0.68_{\pm 0.11}$ & $0.50_{\pm 0.07}$ & $0.03$ & $0.58_{\pm 0.23}$ & $0.67_{\pm 0.15}$
& $16.15_{\pm 4.62}$ & $0.75_{\pm 0.12}$ & $1.45e\text{-}08$ & $0.99$ & $0.69_{\pm 0.23}$ & $0.66_{\pm 0.17}$
& $14.87_{\pm 4.63}$ & $0.72_{\pm 0.12}$ & $0.99_{\pm 0.01}$ & $1.00$ & $0.59_{\pm 0.26}$ & $0.65_{\pm 0.17}$ \\
&                                 & 2
& $16.42_{\pm 4.46}$ & $0.69_{\pm 0.11}$ & $0.50_{\pm 0.08}$ & $0.00$ & $0.65_{\pm 0.20}$ & $0.67_{\pm 0.15}$
& $16.15_{\pm 4.81}$ & $0.75_{\pm 0.11}$ & $2.85e\text{-}10$ & $1.00$ & $0.61_{\pm 0.26}$ & $0.66_{\pm 0.17}$
& $15.16_{\pm 4.23}$ & $0.73_{\pm 0.11}$ & $0.99_{\pm 0.01}$ & $1.00$ & $0.64_{\pm 0.27}$ & $0.65_{\pm 0.14}$ \\
&                                 & 3
& $16.26_{\pm 4.56}$ & $0.69_{\pm 0.11}$ & $0.50_{\pm 0.08}$ & $0.06$ & $0.61_{\pm 0.24}$ & $0.65_{\pm 0.18}$
& $15.83_{\pm 4.70}$ & $0.75_{\pm 0.12}$ & $3.88e\text{-}07$ & $0.94$ & $0.62_{\pm 0.26}$ & $0.66_{\pm 0.20}$
& $15.14_{\pm 4.33}$ & $0.73_{\pm 0.12}$ & $0.99_{\pm 0.01}$ & $1.00$ & $0.67_{\pm 0.25}$ & $0.68_{\pm 0.17}$ \\
&                                 & 4
& $16.26_{\pm 4.52}$ & $0.69_{\pm 0.11}$ & $0.49_{\pm 0.08}$ & $0.03$ & $0.65_{\pm 0.18}$ & $0.70_{\pm 0.15}$
& $16.41_{\pm 4.52}$ & $0.76_{\pm 0.11}$ & $8.00e\text{-}09$ & $0.98$ & $0.67_{\pm 0.24}$ & $0.66_{\pm 0.18}$
& $15.22_{\pm 4.25}$ & $0.74_{\pm 0.12}$ & $0.99_{\pm 0.01}$ & $1.00$ & $0.60_{\pm 0.26}$ & $0.66_{\pm 0.17}$ \\
&                                 & 5
& $16.30_{\pm 4.60}$ & $0.69_{\pm 0.11}$ & $0.50_{\pm 0.06}$ & $0.03$ & $0.60_{\pm 0.23}$ & $0.66_{\pm 0.17}$
& $15.87_{\pm 4.69}$ & $0.75_{\pm 0.12}$ & $3.06e\text{-}08$ & $0.98$ & $0.65_{\pm 0.21}$ & $0.64_{\pm 0.18}$
& $15.01_{\pm 4.41}$ & $0.73_{\pm 0.12}$ & $0.99_{\pm 0.01}$ & $1.00$ & $0.62_{\pm 0.24}$ & $0.64_{\pm 0.17}$ \\
\bottomrule
\end{tabular}
}
\end{sidewaystable}

\clearpage

\begin{figure}
  \centering

  \begin{subfigure}[b]{0.32\linewidth}
    \centering
    \includegraphics[width=\linewidth]{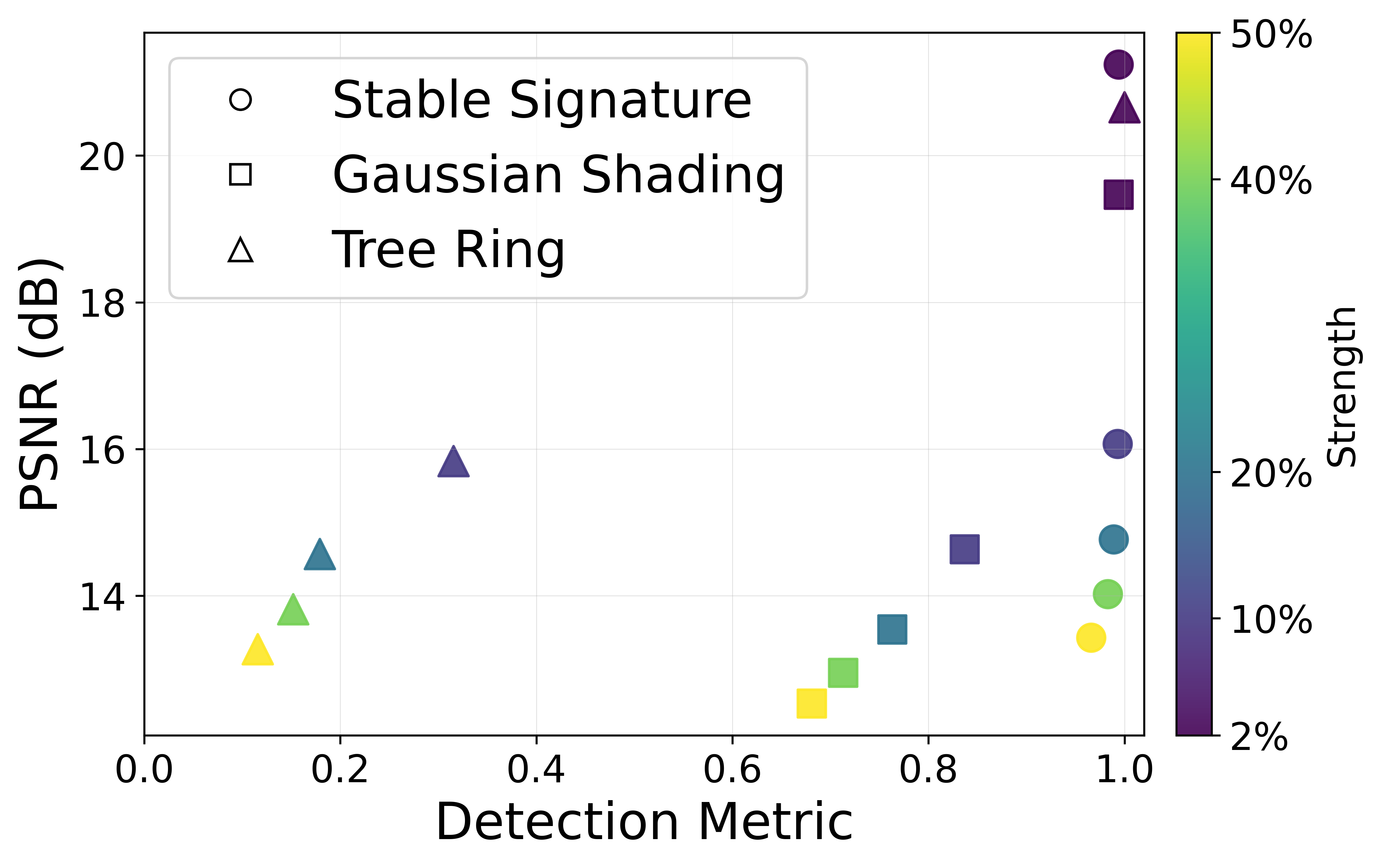}
    \subcaption{Seam Carving}
  \end{subfigure}\hfill
  \begin{subfigure}[b]{0.32\linewidth}
    \centering
    \includegraphics[width=\linewidth]{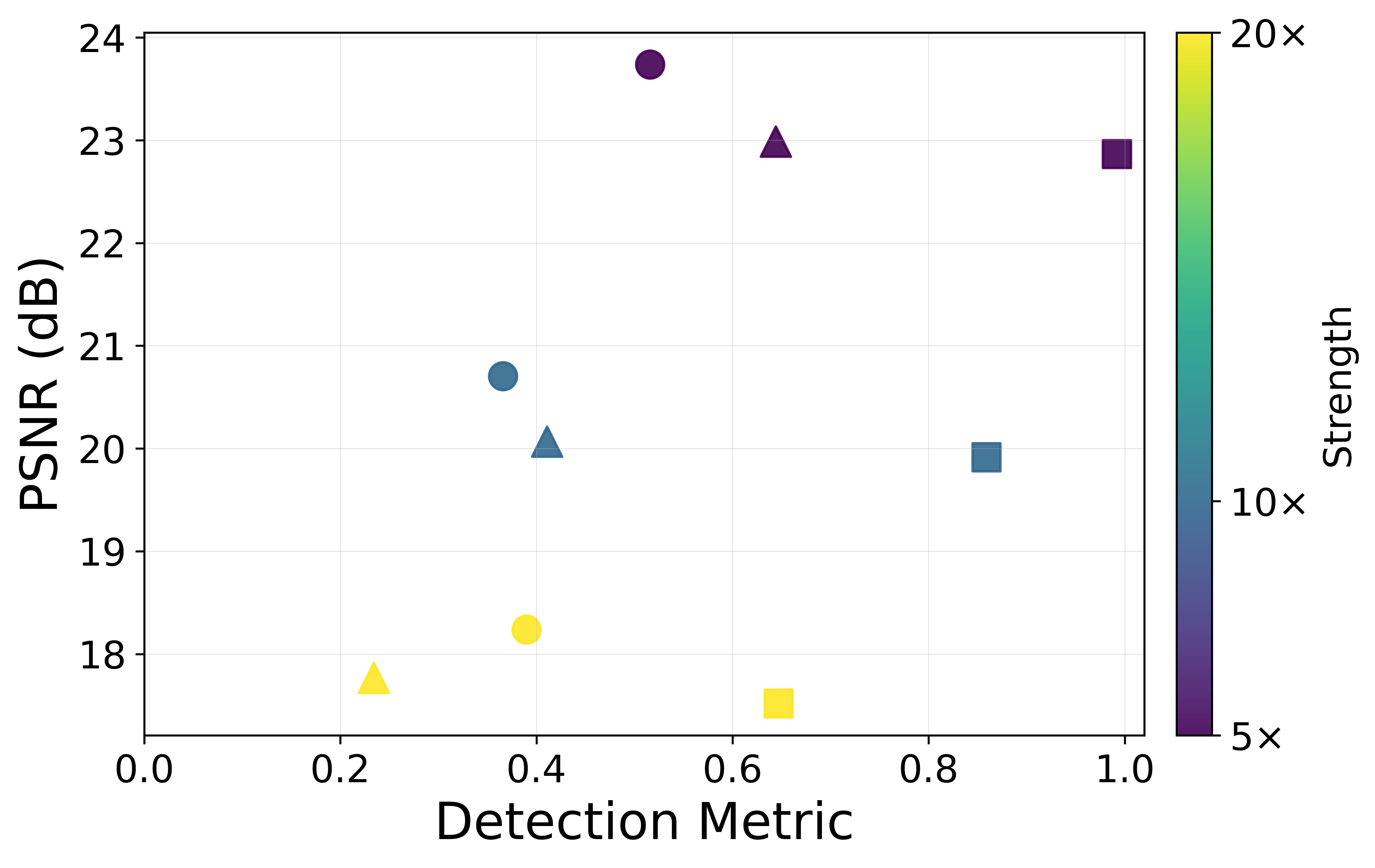}
    \subcaption{Downsampling}
  \end{subfigure}\hfill
  \begin{subfigure}[b]{0.32\linewidth}
    \centering
    \includegraphics[width=\linewidth]{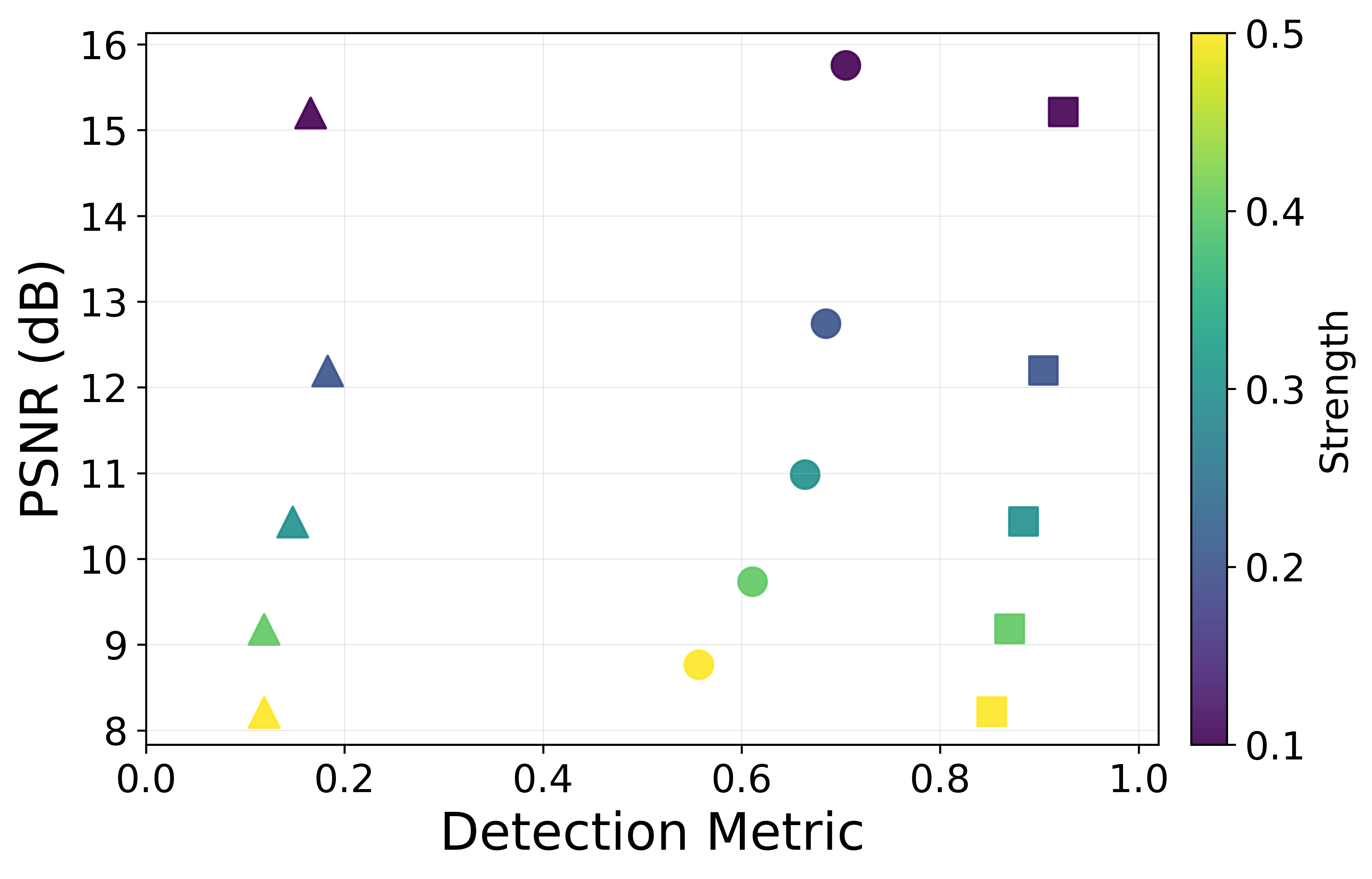}
    \subcaption{Impulse Noise} 
  \end{subfigure}

  \begin{subfigure}[b]{0.32\linewidth}
    \centering
    \includegraphics[width=\linewidth]{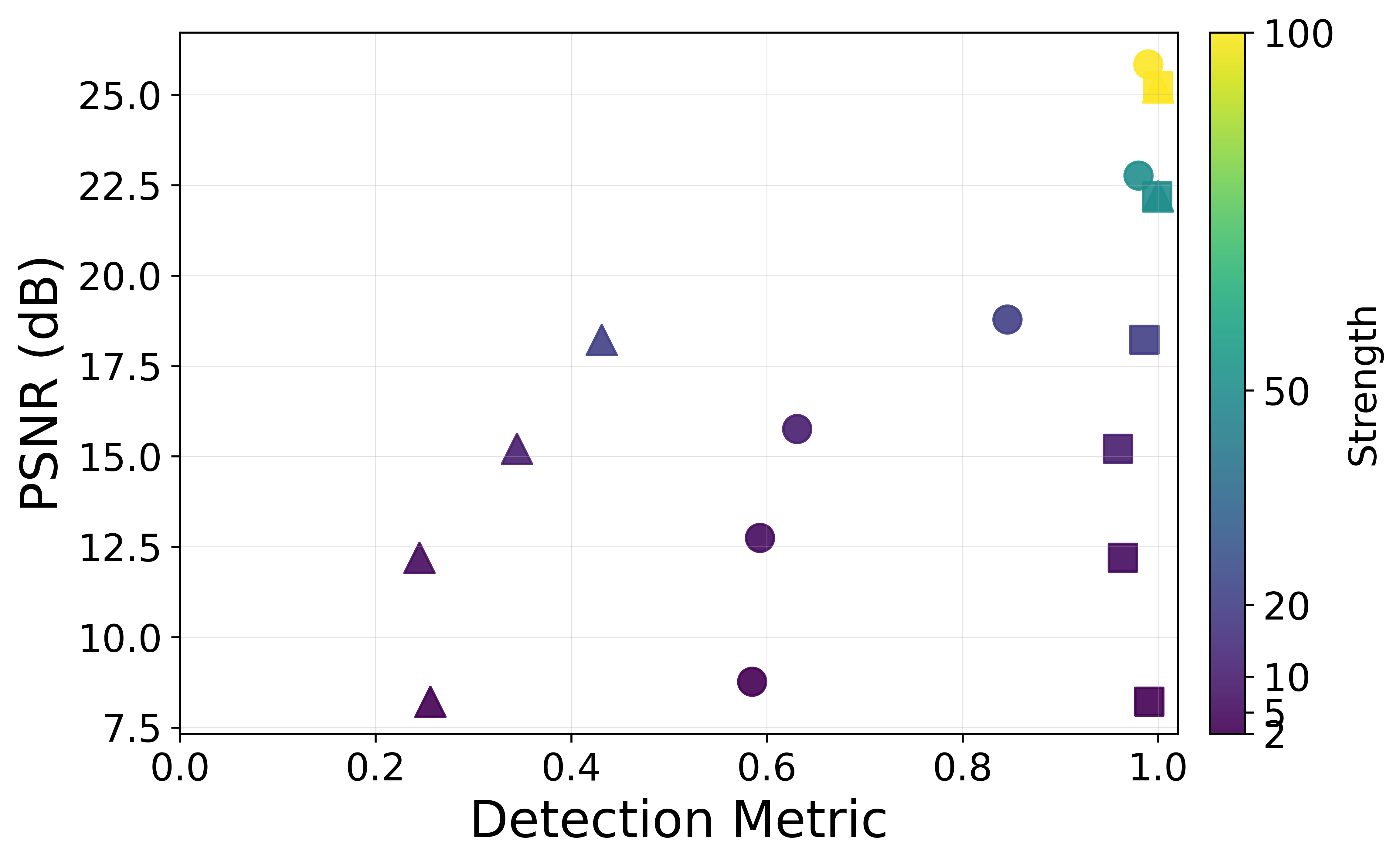}
    \subcaption{Interleaving}
  \end{subfigure}\hfill
  \begin{subfigure}[b]{0.32\linewidth}
    \centering
    \includegraphics[width=\linewidth]{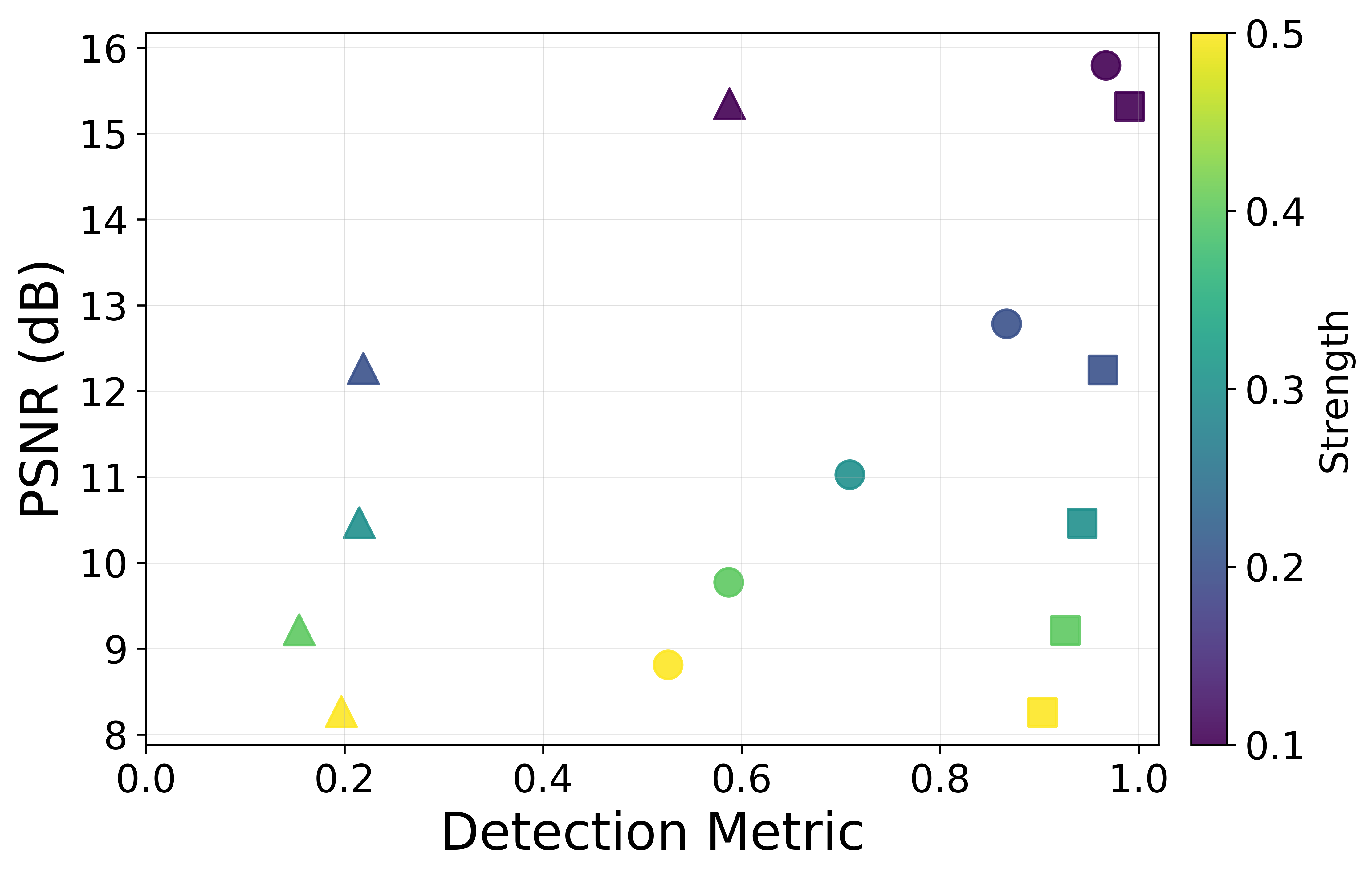}
    \subcaption{Occlusion}
  \end{subfigure}\hfill
  \begin{subfigure}[b]{0.32\linewidth}
    \centering
    \includegraphics[width=\linewidth]{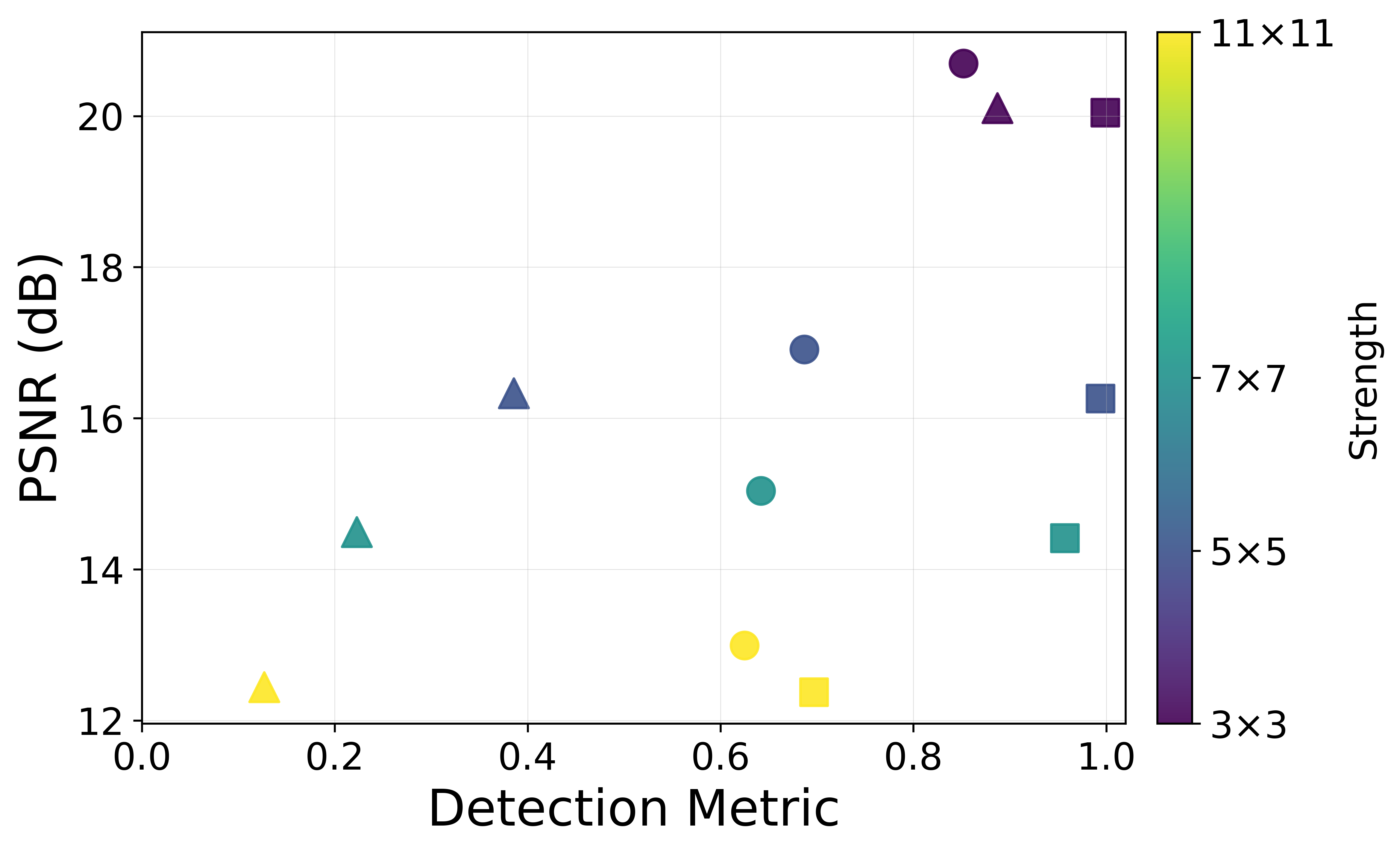}
    \subcaption{Morph. Erosion}
  \end{subfigure}

  \begin{subfigure}[b]{0.32\linewidth}
    \centering
    \includegraphics[width=\linewidth]{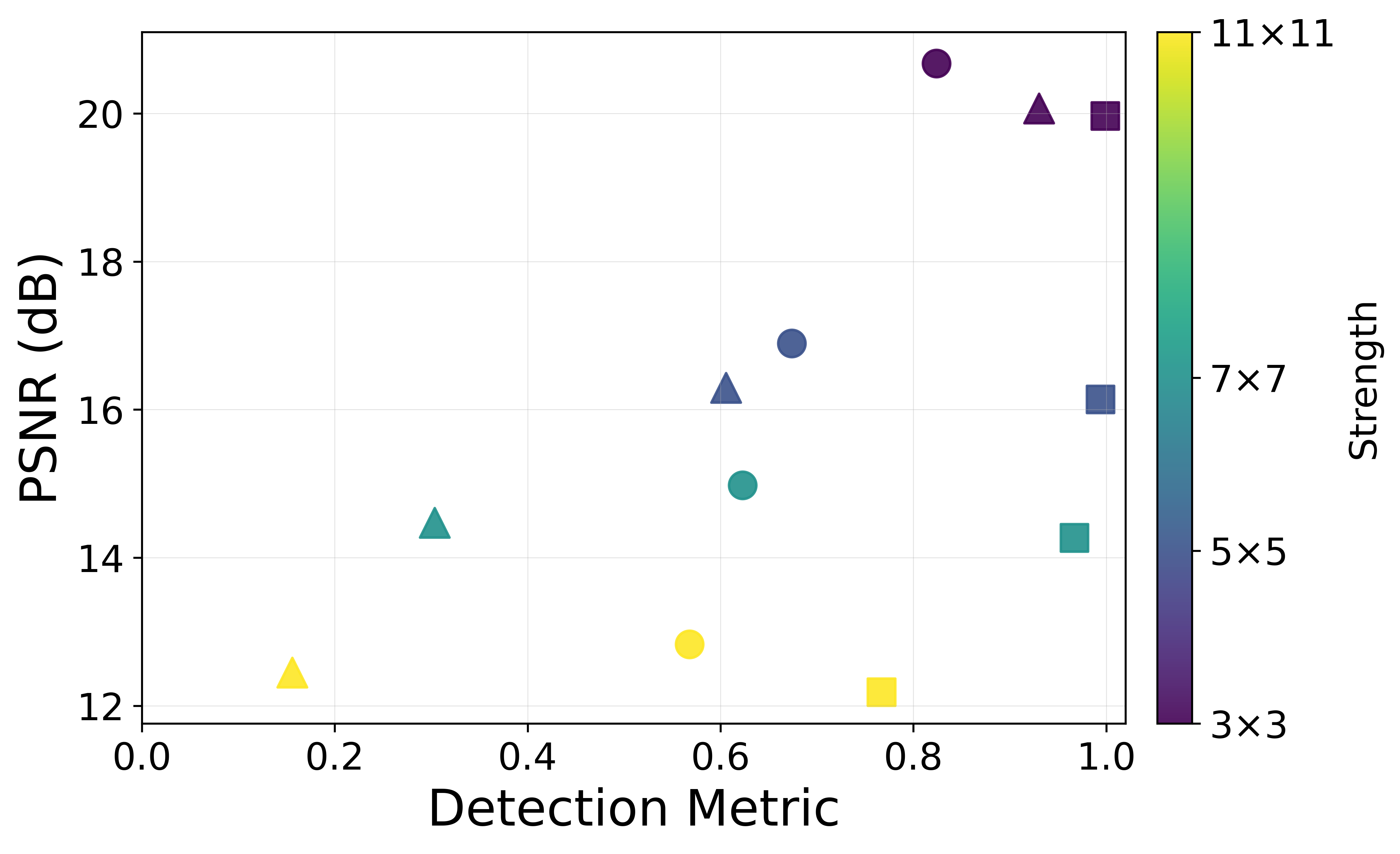}
    \subcaption{Morph. Dilation}
  \end{subfigure}\hfill
  \begin{subfigure}[b]{0.32\linewidth}
    \centering
    \includegraphics[width=\linewidth]{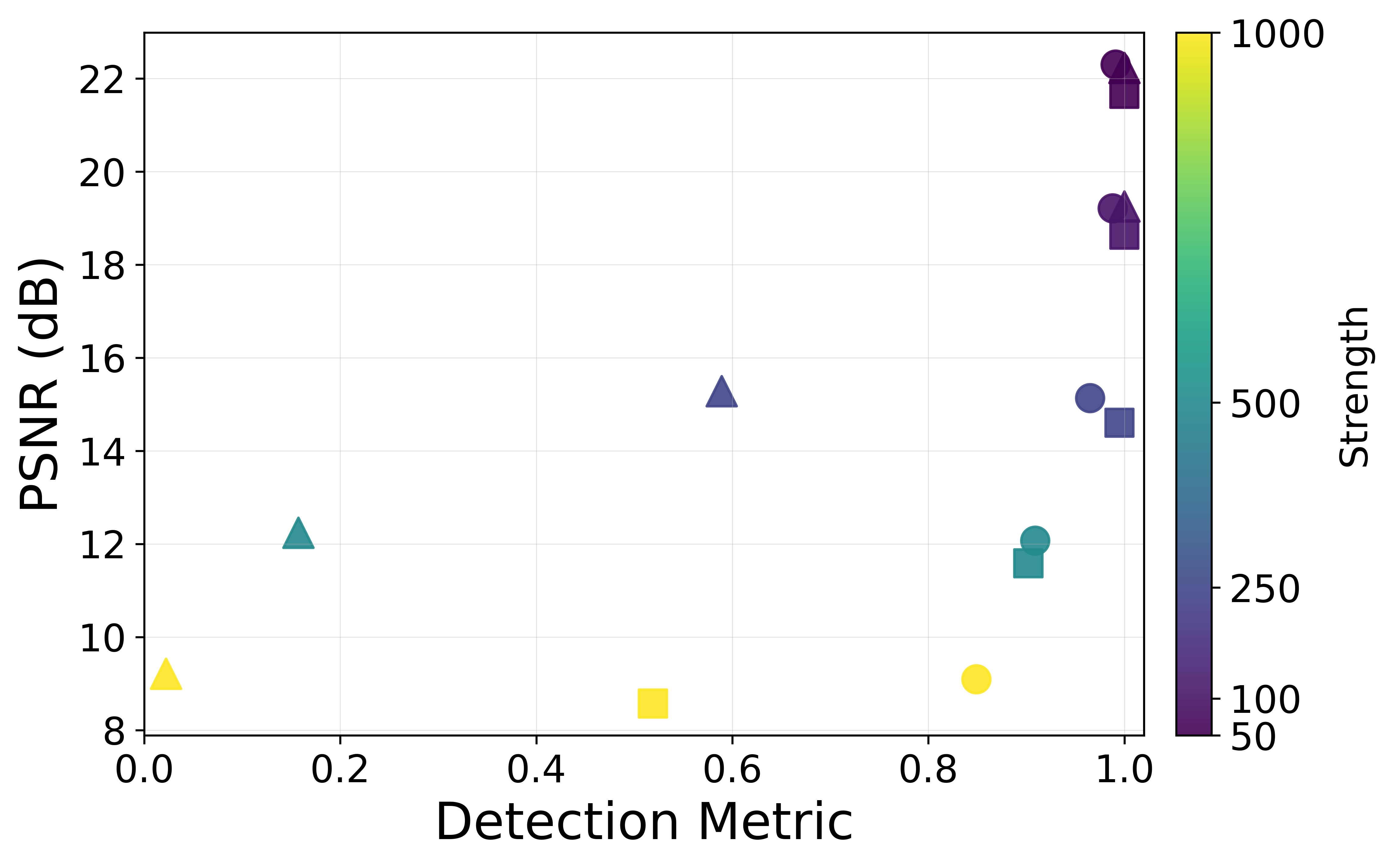}
    \subcaption{Partial Block Shuffling}
  \end{subfigure}\hfill
  \begin{subfigure}[b]{0.32\linewidth}
    \centering
    \includegraphics[width=\linewidth]{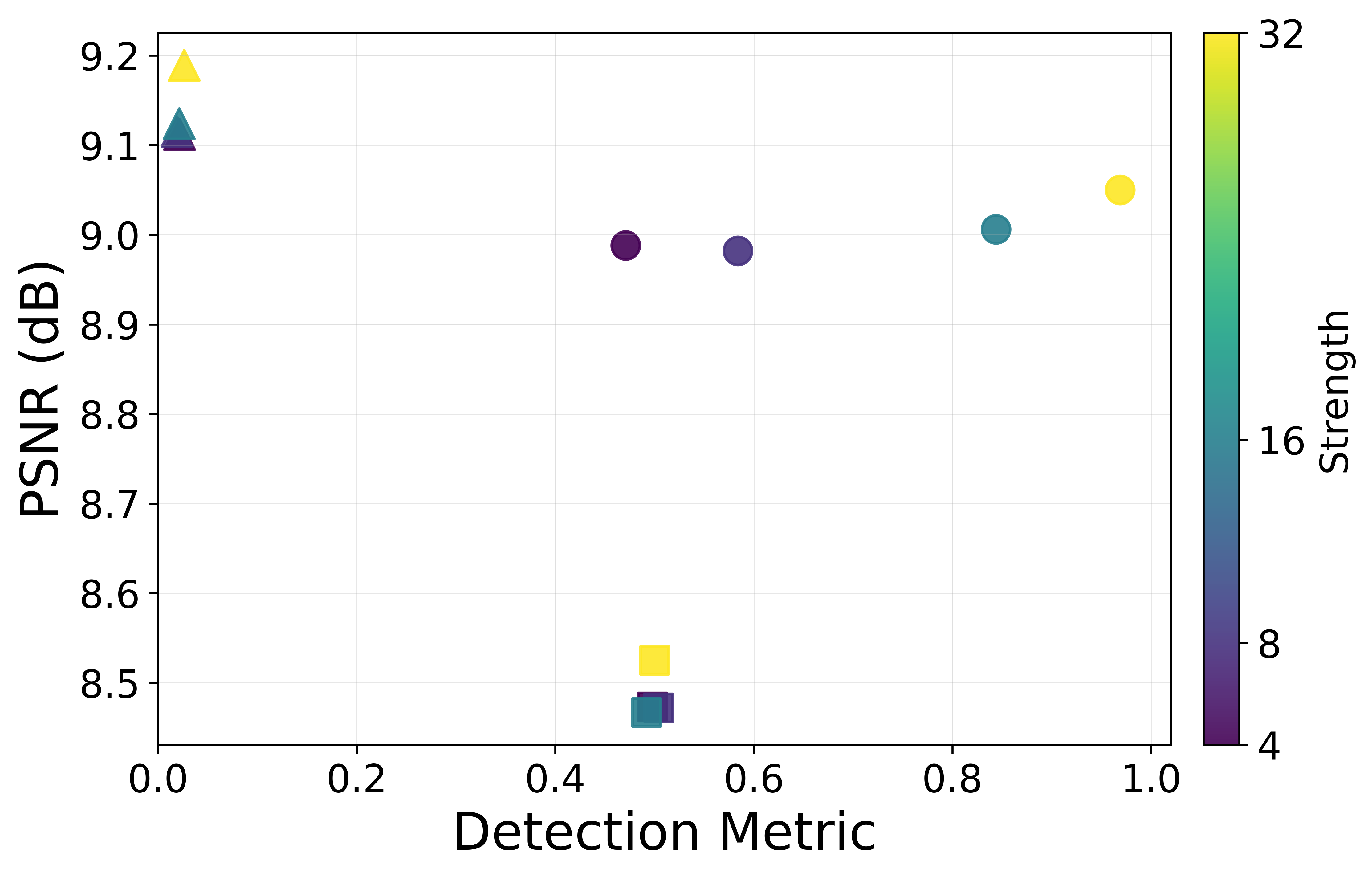}
    \subcaption{Complete Block Shuffling}
  \end{subfigure}
    \vspace{-2mm}
  \caption{Experimental results of watermark detectability vs visual fidelity in PSNR under enhanced image-processing-based perturbations, showing general robustness under these manipulations. Detection metrics are bit accuracy for Stable Signature/Gaussian shading and the normalized negative logarithm of $p$-value for Tree-Ring. Color indicates the perturbation strength, and markers indicate the watermarking schemes.}
  \label{fig:3x3grid_psnr}
\end{figure}

\begin{figure}[!ht]
    \centering
    \includegraphics[width=0.8\linewidth]{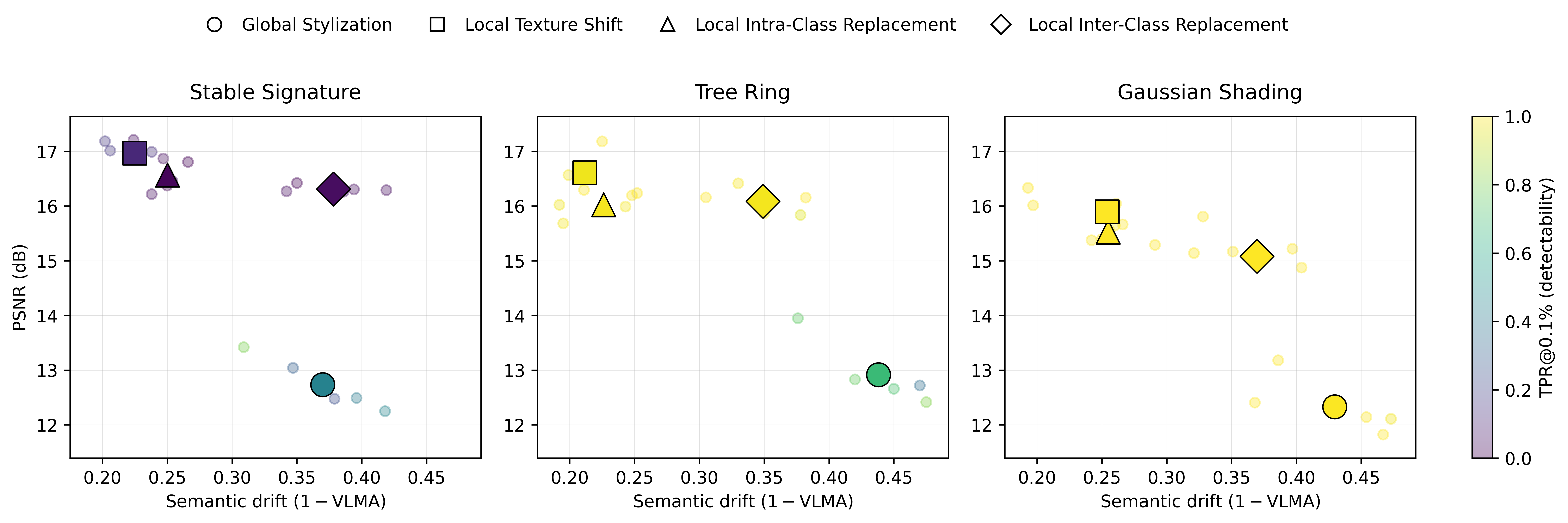}
    \caption{Experimental results of watermark detectability against visual fidelity in PSNR ($y$-axis) and semantic drift ($x$-axis) under semantic perturbations. Color indicates detectability (TPR@0.1\%). Results show watermark detectability can collapse across methods under varying levels of semantic drift, revealing a gap that has not been captured by conventional robustness tests.}
    \label{fig:semantic_tradeoff_psnr}
\end{figure}

\clearpage

\subsection{Qualitative Examples of Semantic Drift Captions}

\begin{figure}[!ht]
    \centering
    \begin{subfigure}[t]{0.33\linewidth}
        \centering
        \includegraphics[width=\linewidth]{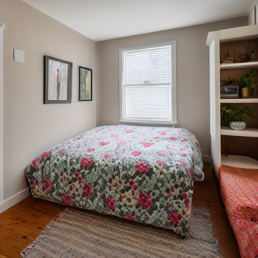}
        \caption{}
    \end{subfigure}\hfill
    \begin{subfigure}[t]{0.33\linewidth}
        \centering
        \includegraphics[width=\linewidth]{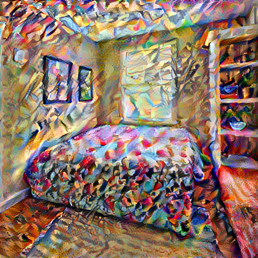}
        \caption{}
    \end{subfigure}
    \begin{subfigure}[t]{0.33\linewidth}
        \centering
        \includegraphics[width=\linewidth]{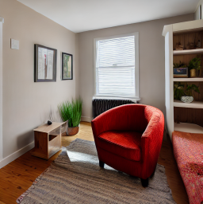}
        \caption{}
    \end{subfigure}
    \caption{(a)~Image watermarked using StableSignature~\cite{fernandez2023stable}, (b)~Global semantic perturbation on (a), (c)~Local inter-object replacement perturbation on (a).}
    \label{fig:blipavsvlma}
\end{figure}

\begin{tcolorbox}[title={BLIP Caption for Fig.~\ref{fig:blipavsvlma}(a)}]
\small
\begin{verbatim}
    there is a bed with a floral comforter in a bedroom, 
    new england architecture, photo taken in 2018, 
    shot on anamorphic lenses, real estate photography, 
    full width, hdr photo, trending on textures. com, 
    by Amelia Robertson Hill, small room
\end{verbatim}
\end{tcolorbox}

\begin{tcolorbox}[title={VLM Caption for Fig.~\ref{fig:blipavsvlma}(a)}]
\small
\begin{verbatim}
    A cozy bedroom featuring a comfortable bed draped in a 
    pink floral bedspread, adorned with a floral comforter 
    and two pillows. The room exudes a warm and inviting 
    atmosphere, with a rug on the floor, a small throw rug, 
    and a rug on the bed. The walls are adorned with 
    pictures and a painting, while a shelf in the corner 
    provides ample storage space. The ceiling is white, 
    and the window is framed by white blinds.
    The bedspread and pillows are white with pink flowers, 
    and the rug on the bed is striped. The room also includes 
    a twin-sized bed, a full-size bed, and a daybed.
\end{verbatim}
\end{tcolorbox}

\begin{tcolorbox}[title={BLIP Caption for Fig.~\ref{fig:blipavsvlma}(b)}]
\small
\begin{verbatim}
    painting of a bedroom with a bed and a bookcase, 
    deepdream, dreamy blurred lens, image dataset, 
    prism lights, 5 0 0 px, calico cat, lacking in 
    three-dimensionality, psychedelic interconnections, 
    sunlight pouring through window, rendered
\end{verbatim}
\end{tcolorbox}

\begin{tcolorbox}[title={VLM Caption for Fig.~\ref{fig:blipavsvlma}(b)}]
\small
\begin{verbatim}
    A cozy room with a comfortable bed and a bookshelf. 
    The bed is made with a multi-colored comforter 
    and blankets. The bookshelf is filled with books 
    and knick-knacks. The wall is yellow and the 
    ceiling is grey. There is a picture hanging on 
    the wall above the bed and a small picture 
    hanging on the wall near the bookshelf. The 
    room is lit by a warm glow from the fireplace.
\end{verbatim}
\end{tcolorbox}

\begin{tcolorbox}[title={BLIP Caption for Fig.~\ref{fig:blipavsvlma}(c)}]
\small
\begin{verbatim}
    there is a red chair in a living room with a book shelf, 
    real estate photography, shot on anamorphic lenses, 
    by Esther Blaikie MacKinnon, narrow hallway, realisitc 
    photo, professional detailed photo, image artifacts, 
    canon 20mm lens, a photorealistic rendering
\end{verbatim}
\end{tcolorbox}

\begin{tcolorbox}[title={VLM Caption for Fig.~\ref{fig:blipavsvlma}(c)}]
\small
\begin{verbatim}
    A red chair sits comfortably in the living room, 
    bathed in the warm glow of the window. 
    The rug beneath the chair is grey, while the 
    armrest and the chair itself are adorned with 
    a red cushion. Pictures hang on the wall, 
    and a white shelf stands proudly in the corner. 
    A green plant flourishes on the floor, and a 
    white light switch provides ample illumination. 
    The floor is made of wood, and the blinds on 
    the window are white. A brown table sits next 
    to the shelf, while a green house plant adds 
    a touch of greenery to the space. The living 
    room is filled with a sense of warmth and comfort.
\end{verbatim}
\end{tcolorbox}

\newpage

\subsection{Quantitative Trend Analysis of Semantic Drift Captions}

\begin{figure}[!h]
    \centering
    \includegraphics[width=0.5\linewidth]{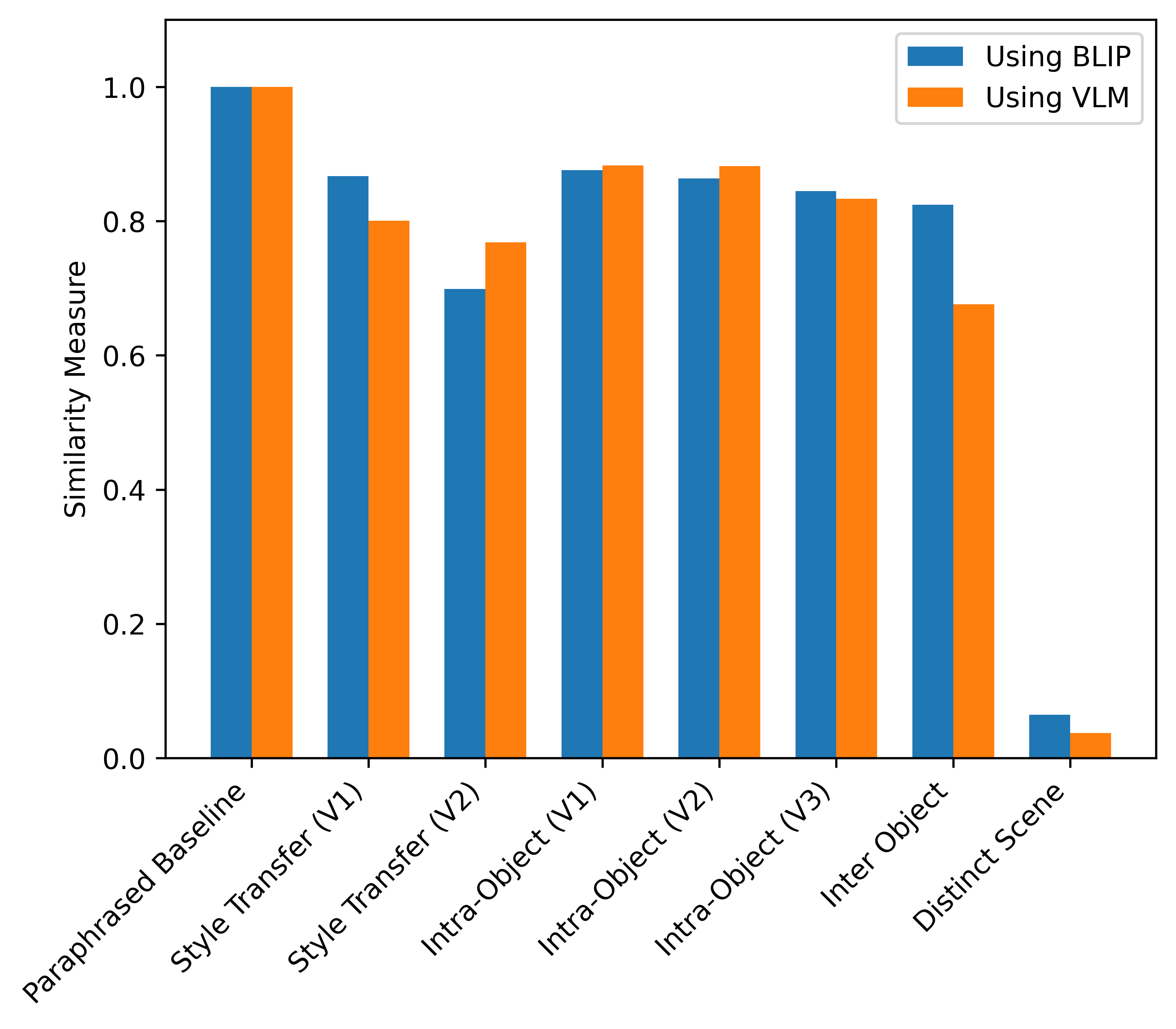}
    \caption{Cosine similarity between unperturbed watermarked image and perturbed watermarked images using BLIP and VLM, including global semantic perturbations with 2 seed variants, intra-object local semantic perturbation with 3 seed variants, inter-object local semantic perturbation, and a distinct scene.}
    \label{fig:bar_blipvlm}
\end{figure}

\EditAni{To enable a fair comparison between BLIP and VLM-based captions, we first post-process BLIP captions using GPT-5 to remove hallucinated or extraneous content. As shown in Fig.~\ref{fig:bar_blipvlm}, BLIP and VLM caption-similarity metrics exhibit consistent qualitative trends. Intra-object local perturbations yield higher similarity than inter-object local perturbations. Although style transfer largely preserves scene semantics, the induced domain shift (such as converting images to an artistic style) reduces caption similarity relative to both local perturbation regimes. Both metrics also correctly assign negligible similarity to unrelated image pairs, indicating strong separation between distinct scenes. Finally, VLM captions provide better discrimination between intra-class and inter-class local perturbations, exhibiting a larger gap in similarity scores than BLIP.}

\begin{table}[!ht]
    \centering
    \caption{(a)~Cosine similarity between unperturbed watermarked image and perturbed watermarked images using BLIP, (b)~Cosine similarity between unperturbed watermarked image and perturbed watermarked images using VLM.}
    \vspace{-5mm}
    \begin{subfigure}[t]{0.49\linewidth}
        \centering
        \includegraphics[trim={3cm 0 0 0},clip, width=\linewidth]{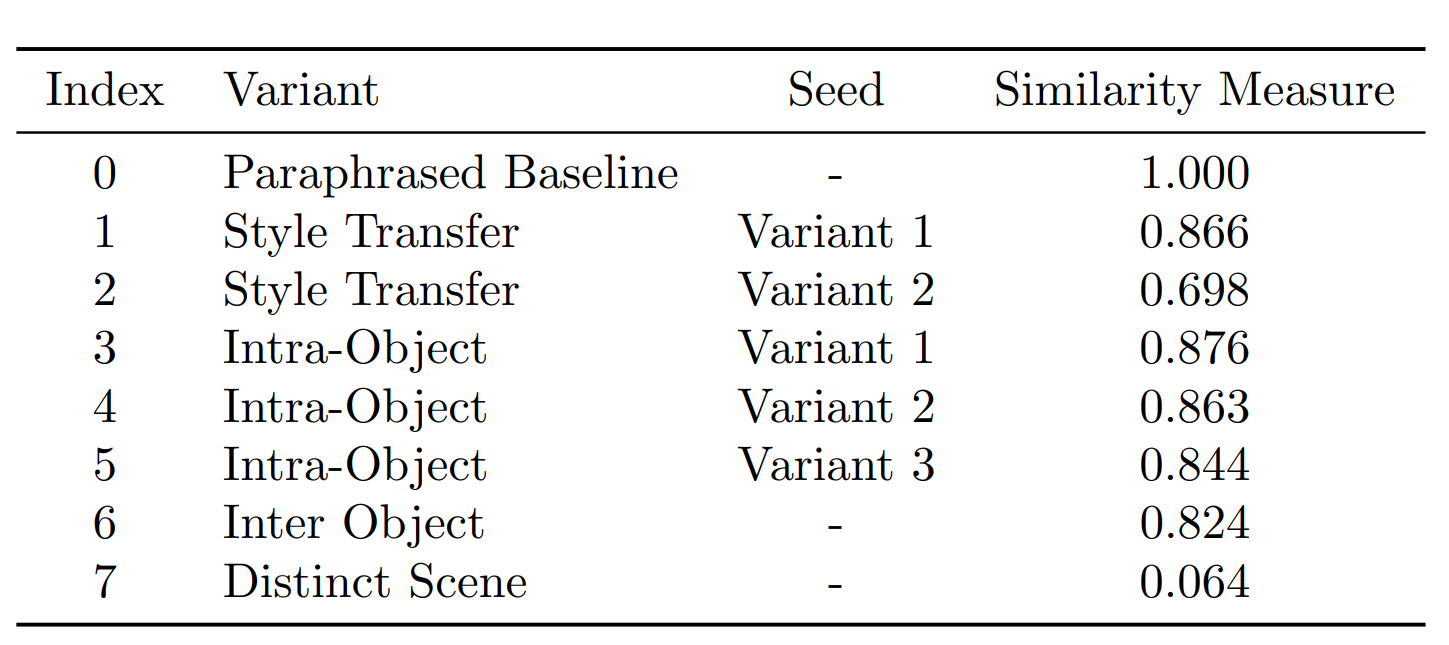}
        \caption{}
    \end{subfigure}
    \begin{subfigure}[t]{0.49\linewidth}
        \centering
        \includegraphics[trim={3cm 0 0 0},clip, width=\linewidth]{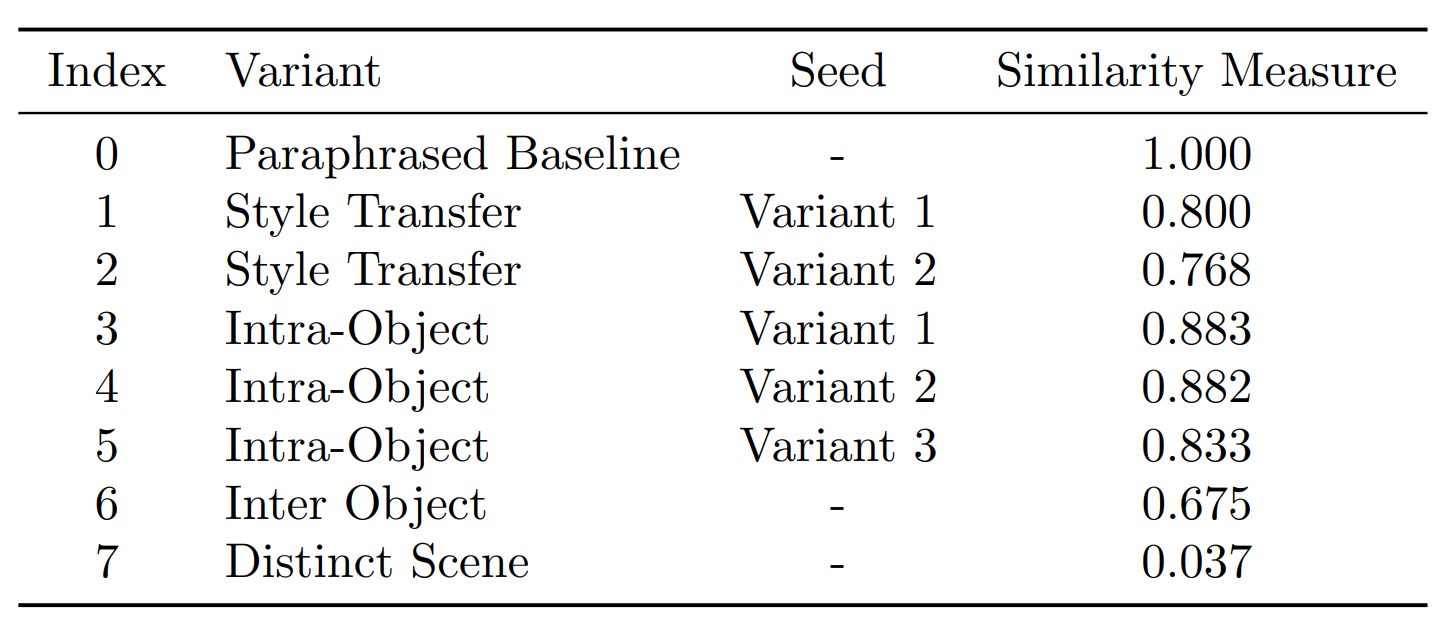}
        \caption{}
    \end{subfigure}
    \label{fig:quant_blipvlm}
\end{table}
\vspace{-15mm}

\clearpage

\subsection{Beyond Captions: Using Scene-Graphs to Quantify Semantic Drift}

\begin{figure}[!h]
    \centering
    \includegraphics[width=0.8\linewidth]{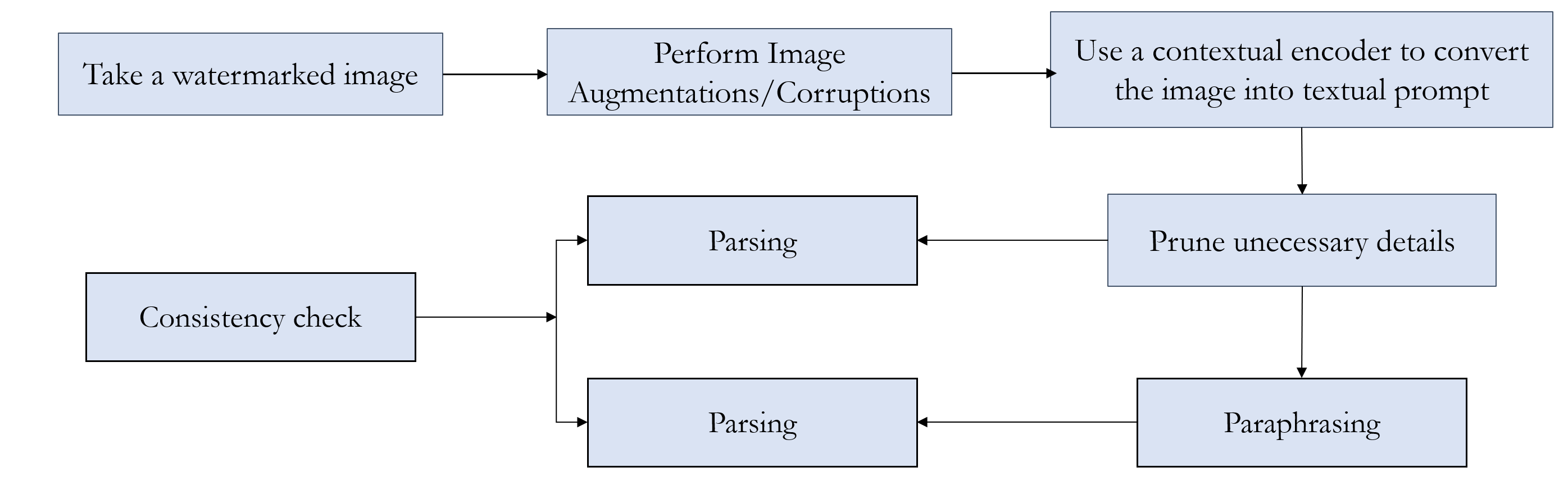}
    \caption{Overview of the scene graph generation (SGG) process.}
    \label{fig:ss}
\end{figure}

\begin{figure}[!h]
    \centering
    \includegraphics[width=0.85\linewidth]{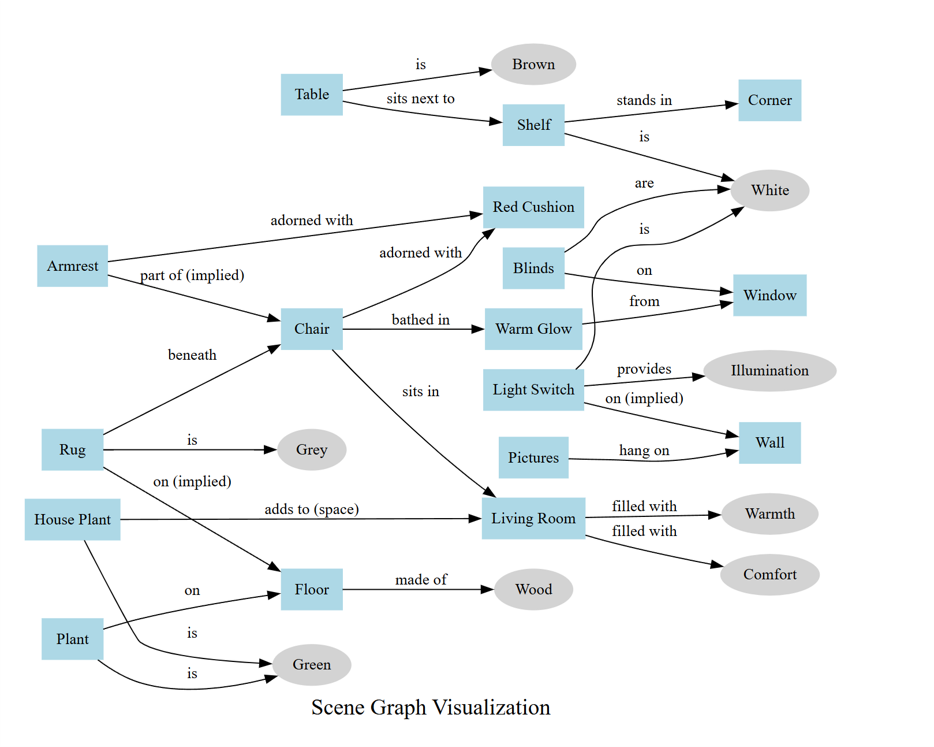}
    \caption{Scene graph of Fig~\ref{fig:blipavsvlma}(a)}
    \label{fig:ss_real}
\end{figure}

\begin{figure}[!ht]
    \centering
    \includegraphics[width=0.7\linewidth]{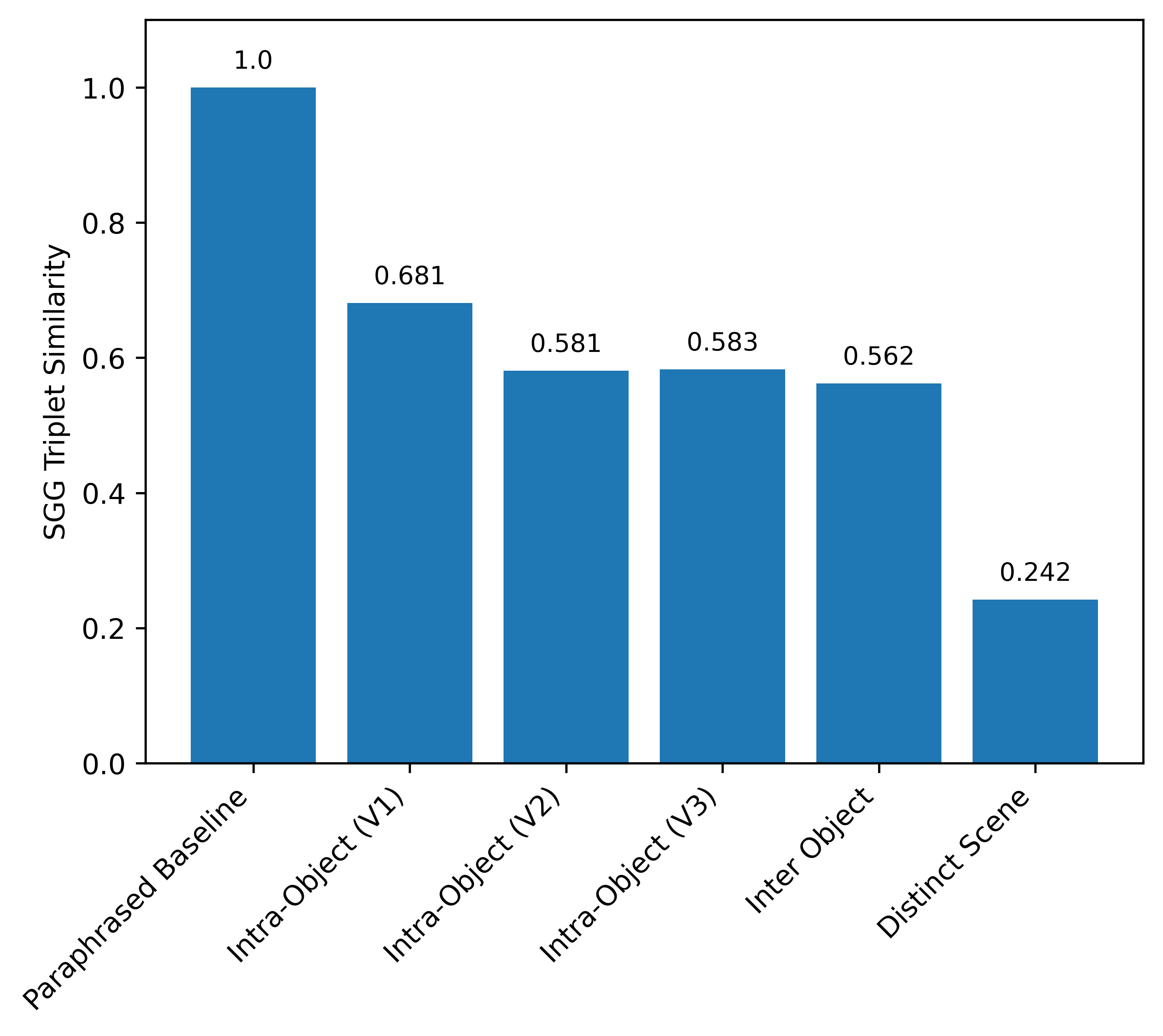}
    \caption{Triplet similarity between unperturbed watermarked image and perturbed watermarked images using SGGs generated by VLM captions, including intra-object local semantic perturbation with 3 seed variants, inter-object local semantic perturbation, and a distinct scene.}
    \label{fig:sgg_blipvlm_bar}
\end{figure}

\EditAni{We exactly follow the scene graph generation pipeline of LLM4SGG~\cite{kim2024llm4sgg} illustrated in Fig.~\ref{fig:ss} to construct scene graphs for both the original and perturbed watermarked images as shown in Fig.~\ref{fig:ss_real}. To quantify similarity between the resulting graphs, we compute a triplet-based alignment score by matching corresponding ⟨subject, predicate, object) relations between the perturbed and original scene graphs. As illustrated in Fig~\ref{fig:sgg_blipvlm_bar}, the trends for triplet similarity align with BLIP and VLM captions.}

\vspace{-10mm}

\subsection{Limitations and Future Work}

\EditAni{Our benchmark enables systematic study of which higher-level image variations different watermarks exploit in the generative setting; however, several limitations remain. First, while we characterize semantic changes induced by our attacks, we do not identify the specific regularities that watermark embedders exploit to yield reliably recoverable signals, and the embedding mechanisms remain largely black-box. Second, our semantic perturbation framework is inherently foreground-centric: the detect-and-replace pipeline prioritizes salient objects and may underrepresent other sources of semantic variation (such as background composition, global geometry, or lighting). Consistent with this, our scene-graph analysis primarily aligns with foreground-object-based analysis. Third, as an initial proof-of-concept work, this paper does not yet provide explanation for the uneven sensitivity of different in-processing methods to syntactic versus semantic perturbations. For instance, Gaussian Shading is highly robust to most semantic edits, yet is more affected by adaptive seam carving than expected. Understanding how watermark schemes can simultaneously exhibit robustness and brittleness---and whether this behavior arises from overfitting to specific latent regularities---remains an open question. Finally, our empirical study focuses on several widely used, representative in-processing methods. Several recent schemes~\cite{chen2025tag,arabi2025seal} are tailored to specific threat models that explicitly incorporate semantic manipulation; assessing their robustness under controlled semantic drift and clarifying their trade-offs is an important direction for future work.

Building on this initial effort, a key direction for future work is to develop explanations for these method-dependent vulnerabilities.
Looking forward, more reliable authentication will likely require multi-layered defenses, including designs that remain detectable under both syntactic and semantic edits. An additional promising direction is watermarking in structured latent spaces to improve interpretability; a central question is how to design such interpretable watermark representations while maintaining robustness to both syntactic and semantic manipulation.}

\subsection{Computing Resources for the Experiments}

\EditAni{All experiments were conducted using three NVIDIA TITAN V GPUs. For Stable Diffusion and Kandinsky, we have used 50 inference steps for diffusion sampling.}

\end{document}